\UseRawInputEncoding
\documentclass[preprint,12pt]{elsarticle}
\usepackage[ruled, vlined, linesnumbered]{algorithm2e}
\usepackage{geometry}
\usepackage{booktabs}
\usepackage{color}
\usepackage{graphicx}
\usepackage{amsmath}
\usepackage{float}
\usepackage{color}
\usepackage{makecell}

\usepackage{amssymb}
\usepackage{dcolumn}
\newcolumntype{d}[1]{D{.}{.}{#1}}

\journal{  }

\begin{document}

\begin{frontmatter}



\title{An Attention-Guided and Wavelet-Constrained Generative Adversarial Network for Infrared and Visible Image Fusion}


\author[a]{Xiaowen Liu}
\author[a]{Renhua Wang}
\author[a]{Hongtao Huo}
\author[a]{Xin Yang}
\author[b]{Jing Li}

\address[a]{School of Information Technology and Cyber Security, People's Public Security University of China, Beijing 100038, China}
\address[b]{School of Information, Central University of Finance and Economics, Beijing 100081, China.}

\begin{abstract}
The GAN-based infrared and visible image fusion methods have gained ever-increasing attention due to its effectiveness and superiority. However, the existing methods adopt the global pixel distribution of source images as the basis for discrimination, which fails to focus on the key modality information. Moreover, the dual-discriminator based methods suffer from the confrontation between the discriminators. To this end, we propose an attention-guided and wavelet-constrained GAN for infrared and visible image fusion (AWFGAN). In this method, two unique discrimination strategies are designed to improve the fusion performance. Specifically, we introduce the spatial attention modules (SAM) into the generator to obtain the spatial attention maps, and then the attention maps are utilized to force the discrimination of infrared images to focus on the target regions. In addition, we extend the discrimination range of visible information to the wavelet subspace, which can force the generator to restore the high-frequency details of visible images. Ablation experiments demonstrate the effectiveness of our method in eliminating the confrontation between discriminators. And the comparison experiments on public datasets demonstrate the effectiveness and superiority of the proposed method.

\end{abstract}




\end{frontmatter}


\section{Introduction}
\label{section:1}
The image fusion technology aims to combine different modality information to generate an informative image \cite{li_pixel-level_2017}. Infrared images can significantly express the high thermal target by capturing the infrared light radiated from the object, but it lacks the texture details \cite{ma_infrared_2019}. Visible images have superior background representation, but it is susceptible to external factors such as weather, light condition and obstacles \cite{ma_infrared_2019}. To this end, we integrate the advantages of infrared and visible modalities to obtain the fusion image, which have both the rich background details and significant target information, and the fusion image can be used for many computer-vision tasks, such as remote sensing \cite{rajah_feature_2018}, video surveillance \cite{chan_fusing_2013,han_fusion_2007}, personnel detection \cite{singh_integrated_2008}, and other fields.

The existing fusion methods include the traditional methods and deep learning-based methods. The traditional methods generally utilize image processing theories to obtain key features from the source images, subsequently construct the fusion result by information fusion and image reconstruction. Typical traditional methods include multiscale transform- \cite{dogra_multi-scale_2017,ADF,FPDE,liu_fusion_2017}, sparse representation- \cite{olshausen_emergence_1996}, subspace- \cite{zhou_principal_2011,kong_technique_2010,cvejic_region-based_2007} and saliency-based \cite{sampat_visible_2015,li_poisson_2019} methods and hybrid methods \cite{bavirisetti_two-scale_2016,liu_general_2015,chen_infrared_2022}. However, the traditional methods require the manual design of activity level measurement and fusion rules \cite{li_pixel-level_2017}. Due to the diversity of infrared and visible modilty information, the design gradually tends to be complex \cite{liu_deep_2018}. Compared with traditional algorithms, the deep learning-based methods have better information representation ability. In \cite{liu_multi-focus_2017}, convolution neural network (CNN) was first introduced into image fusion task. The CNN-based models usually consist of three parts, e.g., encoder, fusion part and decoder. Specifically, the source images are sent into the encoder to extract the typical modality features, and then the extracted features are integrated and fused in the fusion part, subsequently the decoder performs the reconstruction of fusion image.

\begin{figure}[!t]
	\centering
	\includegraphics[width=\textwidth]{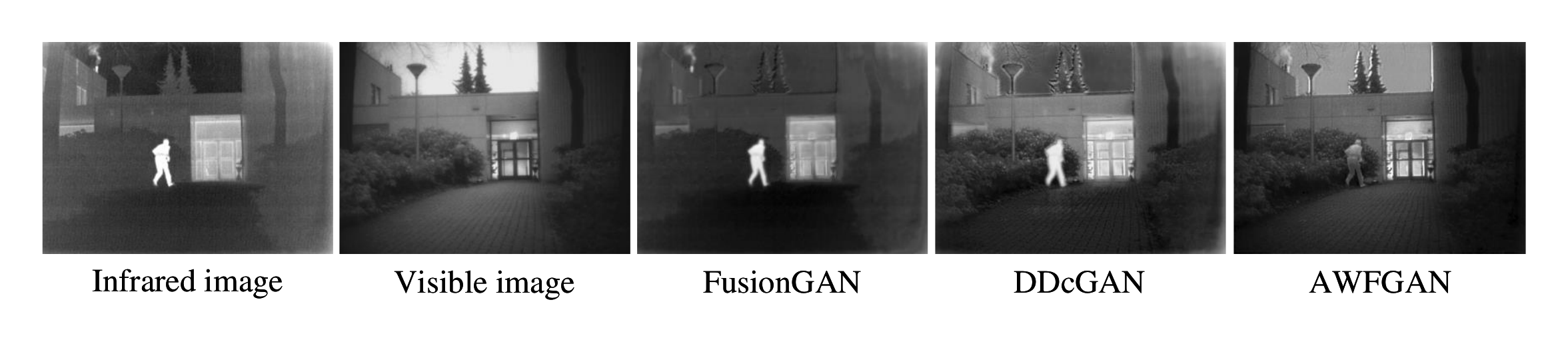}
	\caption{Performance comparison of fusion results. From left to right: infrared image, visible image and the results of FusionGAN, DDcGAN, AWFGAN.}
	\label{fig_1}
\end{figure}

The image fusion can be regarded as the unsupervised task due to the lack of the ground-truth \cite{ma_fusiongan_2019}. The generative adversarial network (GAN) is suitable for solving the unsupervised problem. In \cite{ma_fusiongan_2019}, Ma \emph{et al}. first adopted the adversarial game to enhance the texture details of the fusion images. In their method, a discriminator is used for distinguishing the difference between visible image and the output of the generator. Due to the single discriminator can only focus on one kind of modality information, it might result in the loss of another kind of modality contents. In their subsequent work \cite{ma_ddcgan_2020}, they proposed a dual-discriminator conditional GAN (DDcGAN) to optimize the generator. The DDcGAN adopt the dual-discriminator architecture to force the generator to retain the information from both modalities. Moreover, Li \emph{et al}. proposed a GAN with attention mechanism to achieve better fusion performance \cite{li_attentionfgan_2021}.

The GAN-based fusion methods have made great achievements in the image fusion field, but there are still some limitations. In the GAN-based fusion model, the discriminator is used to distinguish the differences of the global spatial distribution between input images and fused image. Since certain invalid information exists in the source images, such as parts of infrared background contents, the discrimination strategy based on global spatial distribution could lead to the excessive retention of invalid information. As for the methods with dual-discriminator architecture, there is confrontation between the discriminators and the optimizing direction of generator is difficult to be controlled, which might weaken the expression of key modality information.

To solve the above-mentioned problems, we propose a novel GAN-based image fusion method (AWFGAN). In our method, two effective discrimination strategies are designed for the retention of key modality features. The key information of infrared image exists mainly in the target regions, and the contents of infrared background tend to degrade visible texture details in the process of adversarial learning. To solve this problem, we design a spatial domain discrimination strategy with attention guidance. Specifically, in each training iteration, we obtain the spatial attention map from the spatial attention module (SAM) of the generator, and the attention map is utilized to calculating the target mask containing only target regions. Before the discrimination, the target mask is utilized to remove the background contents from the infrared image. Subsequently, the images, which contain only the target content, are sent into the spatial domain discriminator $D_{spa}$ for identification. With the constraint of $D_{spa}$, the generator is forced to focus on the infrared target contents and ignore the infrared background. The visible images have a large amount of texture details, which mainly exists in the high frequency component. To preserve the visible features comprehensively, we adopt the Haar wavelet to decompose the visible and fused images to obtain the low-frequency and high-frequency subimages \cite{heijmans_nonlinear_2000}, subsequently we send the subimages into the frequency domain discriminator $D_{fre}$ for the frequency component-based discrimination. The $D_{fre}$ can adaptively learn the weight of each component and forces the generator to reconstruct the visible details of important frequency bands.

To show the effectiveness of AWFGAN, a schematic comparison is shown in Fig. \ref{fig_1}. Obviously, the result of FusionGAN retains excessive infrared background information. The DDcGAN weakens the interference of infrared content on the background of the fused images, but it still loses a certain visible details. In contrast, our method retains the entire visible background and introduces the visible details into the infrared target.

The main contributions of this paper include the following aspects:
\begin{itemize}
	\item{To weaken the effect of infrared irrelevant information, we propose an attention-guided spatial domain discriminator ($D_{spa}$) to distinguish infrared and fused images, and the attention maps extracted by spatial attention module are utilized to improve the ability of $D_{spa}$, which can force the generator to focus on the infrared target regions.}
	\item{The visible texture details mainly exist in the high frequency component of the visible images. To retain more visible valid information, we design a wavelet-constrained frequency domain discriminator ($D_{fre}$). Specifically, we extend the discrimination range of $D_{fre}$ to the wavelet subspace, therefore, the generator is forced to reconstruct the important high frequency details of the visible image.}
	\item{We compare our method with the state-of-the-art fusion methods by qualitative and quantitative ways. The qualitative and quantitative results demonstrate the superiority of our method on public datasets.}
\end{itemize}

The rest of this paper is organized as follows. Section 2 briefly introduces some related works, including CNN-based methods and GAN-based methods. Section 3 describes the general architecture and the loss functions of the proposed method. The ablation and comparison experiments are presented in the Section 4. Subsequently, Section 5 concludes this paper.

\section{Related Work}
\label{section:2}
The deep learning technology has made great achievement in the image fusion field. Compared with traditional algorithms, deep learning technology has the powerful ability of feature representation. In this section, we review some classical deep learning-based image fusion methods.
\subsection{CNN-based Fusion Methods}
Convolutional neural networks can perceive deep features by a small number of parameters, which is suitable for the tasks with high real-time requirements such as image fusion. 

In \cite{liu_multi-focus_2017}, Liu \emph{et al}. first introduced the CNN into the multi-focus image fusion task. They used a trainable CNN-based model to generate a decision map, and the fusion rule can be adjusted by the decision map. Due to the deep convolution has the risk of information loss, Li \emph{et al}. adopted residual network as the encoder to retain the middle-level features \cite{li_infrared_2019}. In their another work \cite{li_densefuse_2019}, the dense connection was designed to retain more information and a fusion layer is designed to replace the manually designed fusion rules.
The screening of key modality features determines the quality of the fusion image. In \cite{zhang_ifcnn_2020}, Zhang \emph{et al}. proposed a general framework for image fusion. In this method, the encoded features can be selectively retained in different tasks. In order to rationally guide the integration of spatial information, Ma \emph{et al}. used the priori salient target mask to guide the optimization of the model and achieved good results \cite{ma_stdfusionnet_2021}. In \cite{jian_sedrfuse_2021}, Jian \emph{et al}. proposed a symmetric encoder-decoder network to generate the fusion image. In their method, an attention map-based feature fusion strategy was designed to eliminate the interference of redundant information. Meantime, Li \emph{et al}. proposed an fusion framework based on two-stage training \cite{li_rfn-nest_2021}. In the method, the encode-decode part and the fusion part are trained separately by different loss functions. In addition, Li \emph{et al}. combined Transformer with CNN to obtain the contextual and local information of the source image and achieved better results \cite{li_cgtf_2022}.

\subsection{GAN-based Fusion Methods}
The CNN-based methods are generally optimized by the guidance of ground-truth. In the image fusion task, the ground-truth can not be available in advance. On this basis, the CNN-based methods are limited to learning a fusion model to determine the blurring degree of each patch in the source images \cite{ma_fusiongan_2019}. 

To solve above-mentioned problem, Ma \emph{et al}. first introduced the adversarial learning into the image fusion task \cite{ma_fusiongan_2019}. In their method, a discriminator was designed to force the generator to retain more texture details, but the fusion image was trained to be similar with only one kind of the modality information. In their subsequent work \cite{ma_ddcgan_2020}, they proposed a novel GAN (DDcGAN) for multi-modality image fusion, which adopted different discriminators respectively to distinguish the generated image from the source images. The DDcGAN has the defect of inconsistent training. Li \emph{et al}. employed the GAN with Wasserstein distance to build a fusion framework and improved the fusion performance \cite{li_infrared_2020}. In addition, Ma \emph{et al}. proposed a fusion network with multi-classification constraints (GANMcC) for the balanced retention of two types of modality information \cite{ma_ganmcc_2021}. To better express the important contents, Li \emph{et al}. integrated the attention modules into the generator to get the multi-grained attention maps which can enhance the expression of key information \cite{li_multigrained_2021}. In their another work \cite{li_attentionfgan_2021}, the attention modules were also integrated into the discriminators to improve the feature discrimination ability. Moreover, Yang \emph{et al}. combined the traditional method with a dual discriminator network to improve the base and detailed information of the fusion results by integrating guided filtering into the generator \cite{YANG2022116905}.

The GAN-based fusion method has certain drawbacks. On the one hand, since certain invalid information exists in the source images, the discrimination strategy based on global spatial distribution could result in the excessive retention of invalid information. In addition, the methods with dual-discriminator architecture have a confrontation between the discriminators, thus the optimizing direction of the generator is difficult to be controlled, which might weaken the expression of key modality information. Therefore, we propose a new GAN-based fusion method (AWFGAN). Our approach employs two complementary discrimination strategies for improving the optimization process of GAN. Meanwhile, a spatial attention module is used to force the generator and discriminator to focus on infrared targets, and the wavelet transform is utilized to optimize the reconstruction of frequency domain information from visible images.

\begin{figure}[!t]
	\centering
	\includegraphics[width=\textwidth]{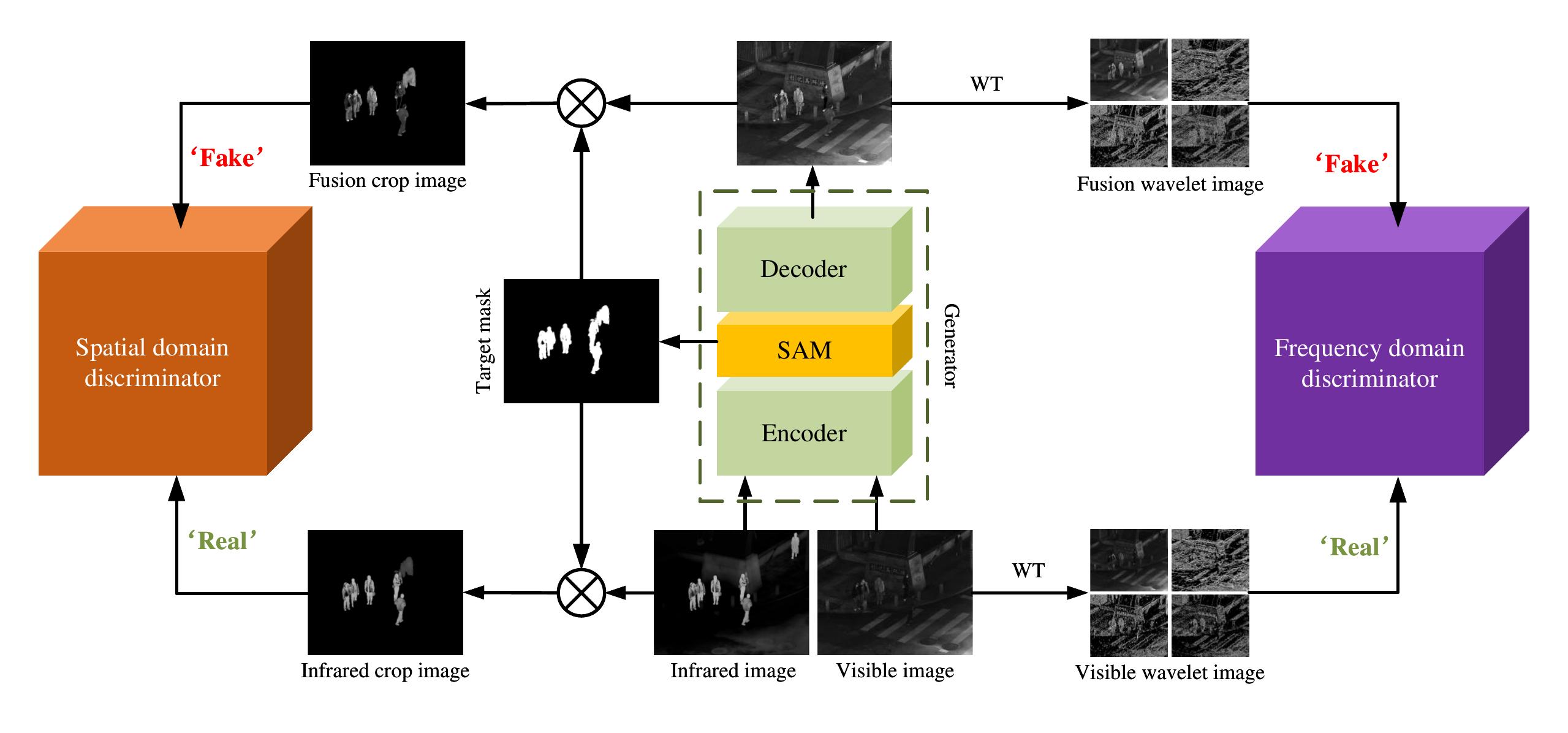}
	\caption{Framework of proposed AWFGAN. The WT denotes the wavelet transform.  $\otimes$ stands for the element-wise multiplication. SAM represents the spatial attention module.}
	\label{fig:fig_2}
\end{figure}

\section{Proposed Method}
In this section, we first introduce the architecture of AWFGAN, and then introduce the loss functions of the generator and discriminators.

\subsection{Framework Overview}

The architecture of AWFGAN is shown in Fig. \ref{fig:fig_2}, which is composed of a generator and two discriminators. Due to the fusion image is expected to have the significant target and the rich background details, we adopt two kinds of discrimination strategies to force the generator to extract the expected features. Meantime, we employ the structure of GAN with Wasserstein distance to improve the training performance \cite{Gulrajani_improved_2017}. The adversarial game can be discribed as follows:
\begin{equation}
	\mathop{min}\limits_{G} \mathop{max}\limits_{D} L_{W}(D,G)=E_{x\sim p_r}[D(x)]+E_{z\sim p_z}[D(G(z))]+\omega E_{\tilde{x}}[(\Vert \nabla{\tilde{x}}D(\tilde{x})\Vert_2-1)^2],
\end{equation}
where $L_W (D,G)$ represents the optimization objectives of discriminators ($D$) and generator ($G$) based on the Wasserstein distance, and the first two terms after the equation denote the Wasserstein distance estimation and the last term stands for gradient penalty factor, $D(x)$ represents the output of discriminator's identification for $x$ and $G(z)$ stands for the generated data of generator, the $\tilde{x}$ represents the uniformly sampled along straight lines connecting pairs of the generated and real data, and $\omega$ denotes penalty coefficient.

In the training stage, the source images are sent into the generator to obtain the fusion image. With the constraint of content loss $L_{con}$, the generated data could retain main infrared and visible modality information. Subsequently, the fusion image and source images are respectively sent into the discriminators to distinguish the similarity of the modality information.

The infrared images lack the background details, which could weak the representation of visible background in the fusion image. To this end, we utilize the spatial attention maps generated by SAM to constrain the interference of infrared background in the adversarial learning process. Under the constraint of the target mask calculated by attention map, the spatial domain discriminator $D_{spa}$ tends to focus on the target regions and ignore the background regions.

\begin{figure}[!t]
	\centering
	\includegraphics[width=\textwidth]{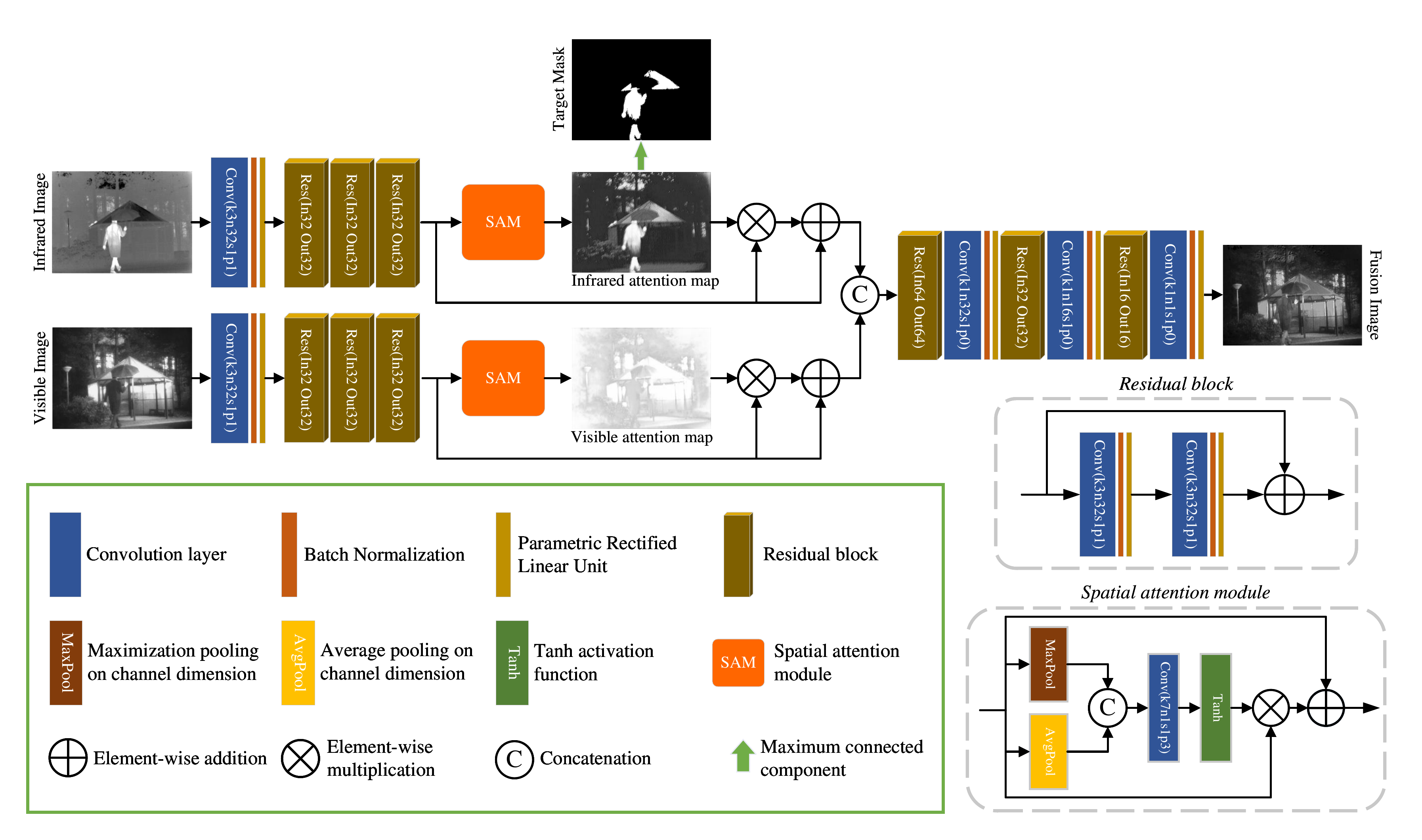}
	\caption{The Framework of generator. The k, n, s, p respectively denote the kernel size, the number of filters, stride, and the padding size. The (In, Out) denote the number of input and output feature maps.}
	\label{fig_3}
\end{figure}

\begin{figure}[htbp]
	\centering
	\includegraphics[scale=0.8]{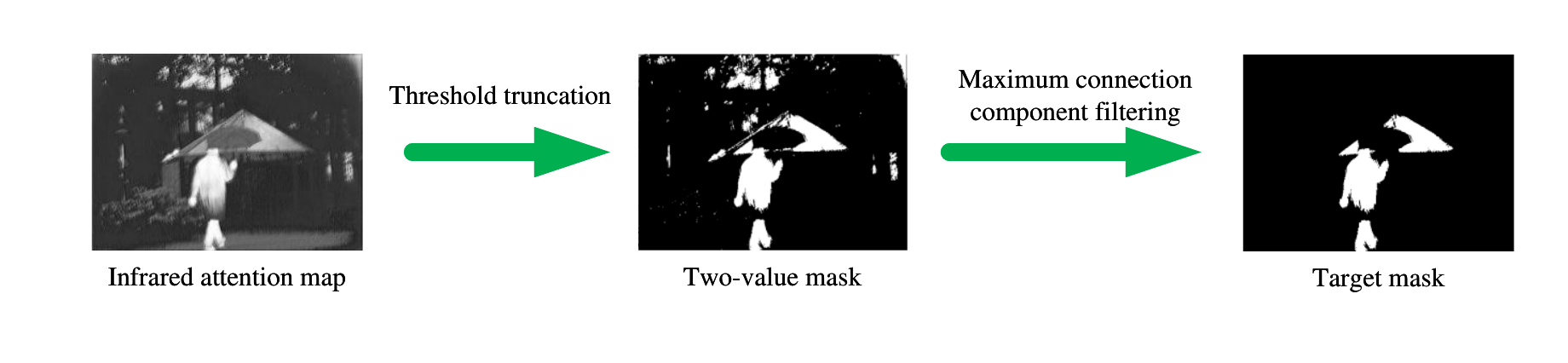}
	\caption{The calculation process of the target mask. Where, the threshold is set to fifty percent of the maximum value of the infrared attention map and three maximum connected components are retained in the target mask.}
	\label{fig_4}
\end{figure}

The visible images contain a large amount of texture details and semantic information, which is difficult to be identified by the distribution of spatial content. Due to the confrontation of the infrared discriminator, the ability of visible discriminator can be susceptible to decay. To solve these problems, we extend the discrimination of visible contents to the wavelet subspace, and a frequency domain discriminator $D_{fre}$ is designed to distinguish the difference between visible image and fusion image in the frequency domain.

\subsection{Generator}

The architecture of generator is shown in Fig. \ref{fig_3}. In the generator, a group of convolutional kernels with $3\times 3$ kernel size are designed to extract the shallow modality features. In addition, three Resblocks are utilized to reinforce the key information. The Resblock is composed of two convolutional layers with $3\times 3$ kernel size, and the skip-connection is used to retain original features, which can alleviate the problem of gradients vanishment/explosion. Moreover, we introduce the SAM into the generator to improve the performance of the key spatial features. The SAM weights the regions by the pooling on the channel dimension and convolution operation, which can generate the attention map by filtering the key regions in the feature maps. 

Subsequently, the extracted modality features are integrated into the feature fusion part, which consists of three Resblocks and three convolutional layers with $1\times 1$ kernel size. Where, the Resblocks are utilized for information fusion, and the $1\times 1$ convolutional kernels are designed to squeeze the channel dimension of feature maps.

To enhance the expression of infrared target, a target mask is calculated by finding the maximum connected component of the infrared attention map. The target mask is used in the subsequent discrimination of $D_{spa}$. The calculation process of the target mask is shown in Fig. \ref{fig_4}.

\begin{figure}[htbp]
	\centering
	\includegraphics[scale=0.75]{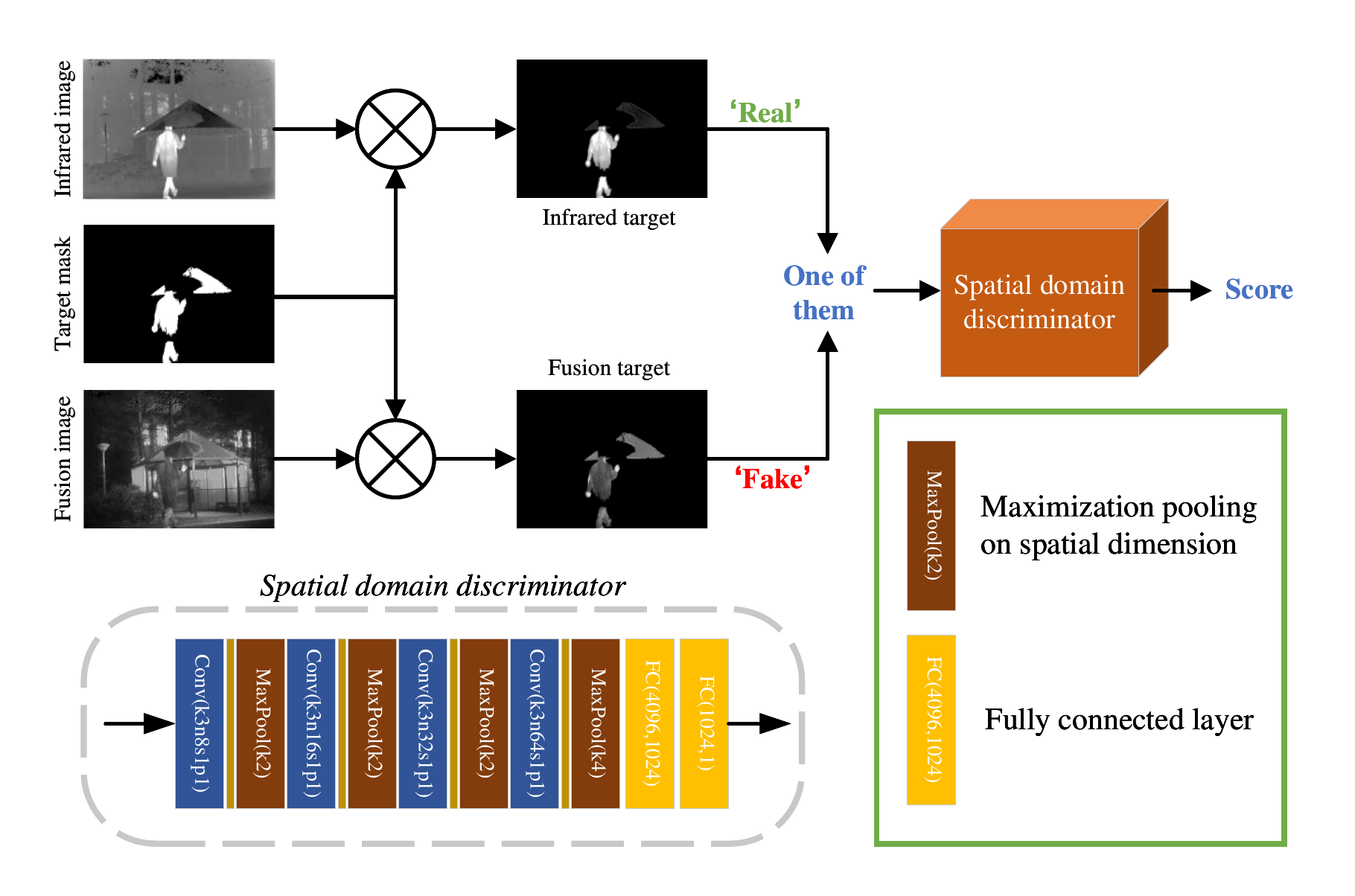}
	\caption{The framework of spatial domain discrimination.}
	\label{fig_5}
\end{figure}

\subsection{The strategy of spatial domain discrimination}

To eliminate the negative impact of infrared information on the fusion background, we design an infrared discrimination strategy with attention guidance, which is shown in Fig. \ref{fig_5}. Specifically, we perform the element-wise multiplication between the target mask and the image to be discriminated and only the target contents are preserved. Subsequently the remaining contents are sent into the spatial domain discriminator $D_{spa}$ to obtain the category score. In the adversarial game, the $D_{spa}$ tends to capture the difference between scores of the results of generator and the infrared images, which force the generator minimize the difference between the two images on the target regions.

The $D_{spa}$ is designed to distinguish the data distribution of target regions between the infrared and fusion images. It is composed of four convolution layers with $3\times 3$ kernel size, maximum pooling operation and two full connection layers. The spatial features are extracted by the convolution kernels. And the pooling operation is used for screening the key features. Finally, the fully connected layers convert the pixel values to a category score by dimensionality reduction.

\subsection{The strategy of frequency domain discrimination}

In order to preserve sufficient visible background, we design a frequency domain discrimination strategy with wavelet transform. Specifically, the image to be discriminated is decomposed to a baseband sub image and three high frequency sub images. We concatenate the sub images and sent them into the frequency domain discriminator $D_{fre}$ to calculate the category score.

\begin{figure}[htbp]
	\centering
	\includegraphics[scale=0.65]{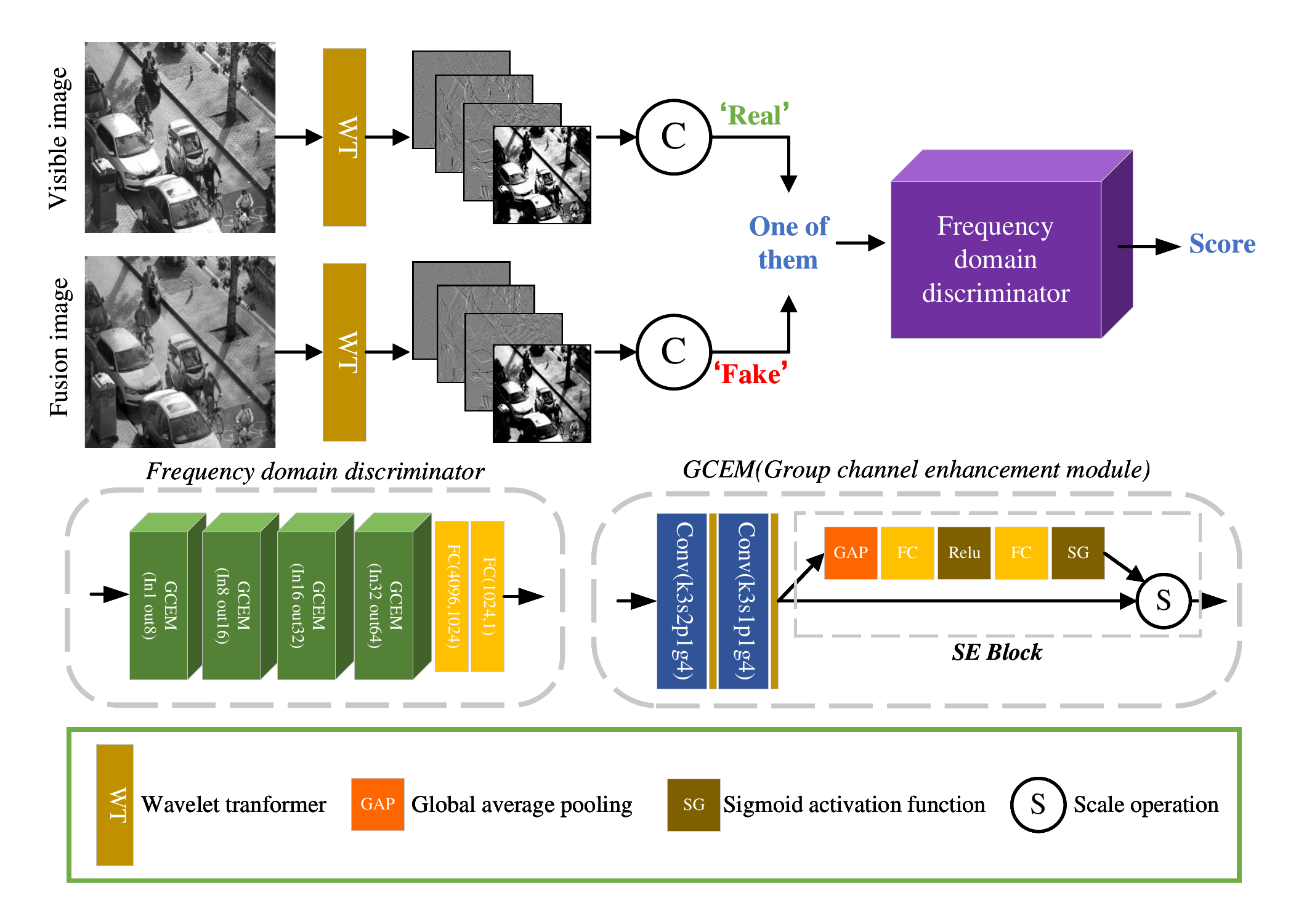}
	\caption{The framework of frequency domain discrimination. The g denotes the group number.}
	\label{fig_6}
\end{figure}

The structure of the $D_{fre}$ is shown in Fig. \ref{fig_6}. To effectively extract the key features in different frequency bands, we design a group channel enhance module (GCEM). The GCEM adopts the group convolution to separately extract the features of the four sub band images. Considering that each frequency component plays a different importance in the expression of the visible background, we utilize the Squeeze-and-Excitation Block (SE Block) to weight the features from different frequency bands. During the discrimination of $D_{fre}$, the background details in the fusion images would be enhanced.

\subsection{Loss Functions}
We build an adversarial game to improve the fusion performance. In the training process, the generator is forced to generate the fusion image which contain both kinds of modality information, and the discriminators are optimized to distinguish the generated data from the source images.

The loss function of generator $L_G$ is composed of two terms, i.e., the content loss $L_{con}$ and the adversarial loss $L_{adv}$:
\begin{equation}
	L_G=L_{adv}+\lambda L_{con} \label{1},
\end{equation}
where the $\lambda$ denotes the weight. The $L_{adv}$ is defined as:
\begin{equation}
	L_{adv}=-E_{z\sim p_g}[D_{spa}(z)]-E_{z\sim p_g}[D_{fre}(z)]\label{2},
\end{equation}
where $p_g$ represents the generative data distribution.
The content loss is used to force the fusion image to preserve basic content of the source images, which is modified as:
\begin{equation}
	L_{con}=MSE(I_{ir},I_f)+\gamma SSIM(I_{vi},I_f)\label{2},
\end{equation}
where the $MSE$ and $SSIM$ denote the mean square error and structural similarity respectively, $\gamma$ is the weight. $MSE$ aims to constrain the pixel difference between infrared image and fusion image and $SSIM$ aims to constrain the structure difference between visible image and fusion image. The $MSE$ and $SSIM$ are respectively defined as follows:
\begin{equation}
	MSE(X,Y)=\frac{1}{HW}\Vert X-Y \Vert_2^2\label{3},
\end{equation}
\begin{equation}
	SSIM(X,Y)=\frac{2\sigma_{XY}+C}{\sigma_X^2+\sigma_Y^2+C},
\end{equation}
where $X$ and $Y$ denote the input images, $H$ and $W$ stand for the length and width of the image, $\Vert{\,}\Vert_2$ represents the $l_2$ norm, $\sigma_{XY}$ stands for the covariance between $X$ and $Y$, $\sigma_X^2$ and $\sigma_Y^2$ are the variance of $X$ and $Y$, $C$ is the constant utilzied to maintain stability.

The loss function of $D_{spa}$ is defined as follows:
\begin{equation}
	L_{D_{spa}}=-E_{x\sim p_{ir_{spa}}}[D_{spa}(x)]+E_{z\sim p_{g_{spa}}}[D_{spa}(z)]+\alpha E_{\tilde{x}}[(\Vert \nabla{\tilde{x}}D_{spa}(\tilde{x})\Vert_2-1)^2],
\end{equation}
where $p_{ir_{spa}}$ and $p_{g_{spa}}$ represent the data distribution of the target region of the infrared image and the generative data respectively, $D_{spa}(x)$ represents the identification result of $x$ by $D_{spa}$, the last term is a gradient penalty factor, and $\alpha$ is a penalty coefficient. 
		
The loss function of $D_{fre}$ is defined as follows:
\begin{equation}
	L_{D_{fre}}=-E_{x\sim p_{vi_{fre}}}[D_{fre}(x)]+E_{z\sim p_{g_{fre}}}[D_{fre}(z)]+\beta E_{\tilde{x}}[(\Vert \nabla{\tilde{x}}D_{fre}(\tilde{x})\Vert_2-1)^2],
\end{equation}
where $p_{vi_{fre}}$ and $p_{g_{fre}}$ represent the data distribution of the frequency  component of the visible image and the generative data respectively, $D_{fre}(x)$ represents the identification result of $x$ by $D_{fre}$, the last term is a gradient penalty factor, and $\beta$ is a penalty coefficient. 

\section{Experimental Results and Analysis}
In this section, we first introduce the experiment details, include the testing and training process. And then we present the ablation experiments to prove the validity of our contributions. Finally, we compare AWFGAN with eight state-of-the-art algorithms and present the qualitative and quantitative results.

\subsection{Experiment Details}
1) Training details: AWFGAN is trained on the LLVIP public dataset. The LLVIP consists of training data and testing data, and the training data have 12025 pairs of images. To preserve the entire scene of the source images, we use the whole image as the input image. Since the discriminator require the fixed input size, we transform the size of training images to $256\times 256$. 
As mentioned above, we adopt the content loss $L_{con}$ and the adversarial loss $L_{adv}$ to optimize the generator. We further summarize the training procedure of AWFGAN in Algorithm 1. All experiments are operated on the NVIDIA Tesla K80 with 12 GB memory and AMD Ryzen 7 4800U with 1.80 GHz.

\begin{algorithm}[!t]
	\caption{Training procedure} \label{Training procedure}
	\begin{small}
		\SetAlgoLined 
		\BlankLine
		\KwIn{Infrared images $I_{ir}$ and visible images $I_{vi}$}
		\KwOut{Fusion image $F$}
		\For{$n$ epochs}{
			\While{$k$ < the number of critic}{
				Select visible images $I_{vi}^i$ and infrared images $I_{ir}^i$\;
				Generate the fusion image $I_f^i$ and infrared attention maps $AM_{ir}^i$ by Generator $G$\;
				Update discriminator $D_{spa}$ by SGDOptimizer to minimize $L_{D_{spa}}$ in Eq. (7)\;
				Obtain the wavelet subimages of fusion images and visible images \{$LL_{f}^i$, $HL_{f}^i$,$LH_{f}^i$,$HH_{f}^i$\} and \{$LL_{vi}^i$, $HL_{vi}^i$,$LH_{vi}^i$,$HH_{vi}^i$\}\;
				Update discriminator $D_{fre}$ by SGDOptimizer to minimize $L_{D_{fre}}$ in Eq. (8)\;			
			}
			Select visible images $I_{vi}^i$ and infrared images $I_{ir}^i$\;			
			Genarate the fusion image $I_f^i$ by Generator $G$\;
			Update Generator $G$ by SGDOptimizer to minimize $L_G$ in Eq. (2)\;			
		}
	\end{small}
\end{algorithm}

2) Testing details: we choose three public datasets as the testing data, \emph{i.e.}, LLVIP, TNO and RoadScene. Among them, the TNO dataset mainly describes the diverse military scene, in which we select 30 pairs of images to perform test experiment. The images of RoadScene and LLVIP dataset mainly describe the complex traffic scenes. In the experiments, the whole RoadScene dataset and 80 pairs of images of LLVIP dataset are selected to test our method.

The fusion performance is evaluated by qualitative and quantitative
manners. The qualitative manner is based on the objective human visual evaluation. For quantitative analysis, six metrics are chosen to evaluate the fusion performance, and the metrics include mutual information (MI) \cite{qu_information_2002}, entropy (EN) \cite{van_aardt_assessment_2008}, standard deviation (SD) \cite{rao_-fibre_1997}, spatial frequency (SF) \cite{eskicioglu_image_1995}, visual information fidelity for fusion (VIFF) \cite{VIFF} and the sum of the correlations of differences (SCD) \cite{SCD}. The values of these metrics are positively correlated with the quality of the fusion image. We briefly introduce these metrics below, where, A, B and F represent the infrared, visible and fusion images respectively, and the size of them is $M\times N$.

MI can measure the amount of both kinds of modality information contained in the fusion image. The formula is as follows:
\begin{equation}
	MI=MI_{A,F}+MI_{B,F},
\end{equation}
\begin{equation}
	MI_{X,F}=\sum_{x,f}P_{X,F}(x,f)\log{\frac{P_{X,F}(x,f)}{P_X(x)P_F(f)}},
\end{equation}
where $MI_{A,F}$, $MI_{B,F}$ denote the amount of infrared modality and visible modality information contained in the fused image respectively, $X$ represents the source images, which is equal to either A or B, $P_X(x)$ and $P_F(f)$ denote the marginal histograms of source image $X$ and fused image $F$ respectively, $P_{X,F}(x,f)$ denotes the joint histogram of source image $X$ and fused image $F$.

EN is an objective evaluation metric that measures the amount of information contained by the image, and the formula is expressed as follows:
\begin{equation}
	EN=-\sum_{a=1}^{L-1} P(a)\log_2P(a),
\end{equation}
where $a$ represents gray levels of fusion image, $L$ represents the number of gray levels and $P(a)$ is the normalized histogram of the corresponding gray level.

SD reflects the distribution and contrast of the image, which is based on the statistical concept. The mathematical definition is as follows:
\begin{equation}
	SD=\frac{1}{MN}\sqrt{\sum_{i=1}^{M} \sum_{j=1}^{N} (F(i,j)-\overline{F})},
\end{equation}
where $\overline{F}$ denotes the mean value of the fused image.

SF calculates the row frequency and column frequency and evaluates the texture details of the image from the gradient distribution. The mathematical definition is as follows:
\begin{equation}
	SF=\sqrt{RF^2+CF^2},
\end{equation}
where $RF$ represents the row frequency and $CF$ represents the column frequency. The formulae of $RF$ and $CF$ are expressed as follows:
\begin{equation}
	RF=\sqrt{\sum_{i=1}^{M} \sum_{j=1}^{N} (F(i,j)-F(i,j-1))},
\end{equation}
\begin{equation}
	CF=\sqrt{\sum_{i=1}^{M} \sum_{j=1}^{N} (F(i,j)-F(i-1,j))}.
\end{equation}

VIFF is consistent with the human visual system and donates the information fidelity of the fused image. In the processing of calculation, the source images and fused image are firstly filtered and divided into different blocks. Subsequently, each block is evaluated the distortion level of the visual information. Finally, the VIFF of each sub-band is calculated and aggregated into a total VIFF.

SCD can reasonably evaluate the fusion quality by calculating the correlation between the different information computed by employing the source images and the fused image. The formula is as follows:
\begin{equation}
	SCD=r(A,F)+r(B,F),
\end{equation}
where $r(.)$ represents the correlation function, and the formula of it is as follows:
\begin{equation}
	r(X,F)=\frac{\sum_{i} \sum_{j} (X(i,j)-\bar{X})}{(\sqrt{(\sum_{i} \sum_{j} (X(i,j)-\bar{X}^2)(\sum_{i} \sum_{j} (F(i,j)-\bar{F})^2)}},
\end{equation}
where $X$ represents the source images, and $bar{X}$ and $bar{F}$ are the average of the pixel values of $X$ and $F$.

\subsection{Generator Performance Analysis }

\begin{figure}[H]
	\centering
	\includegraphics[scale=1.00]{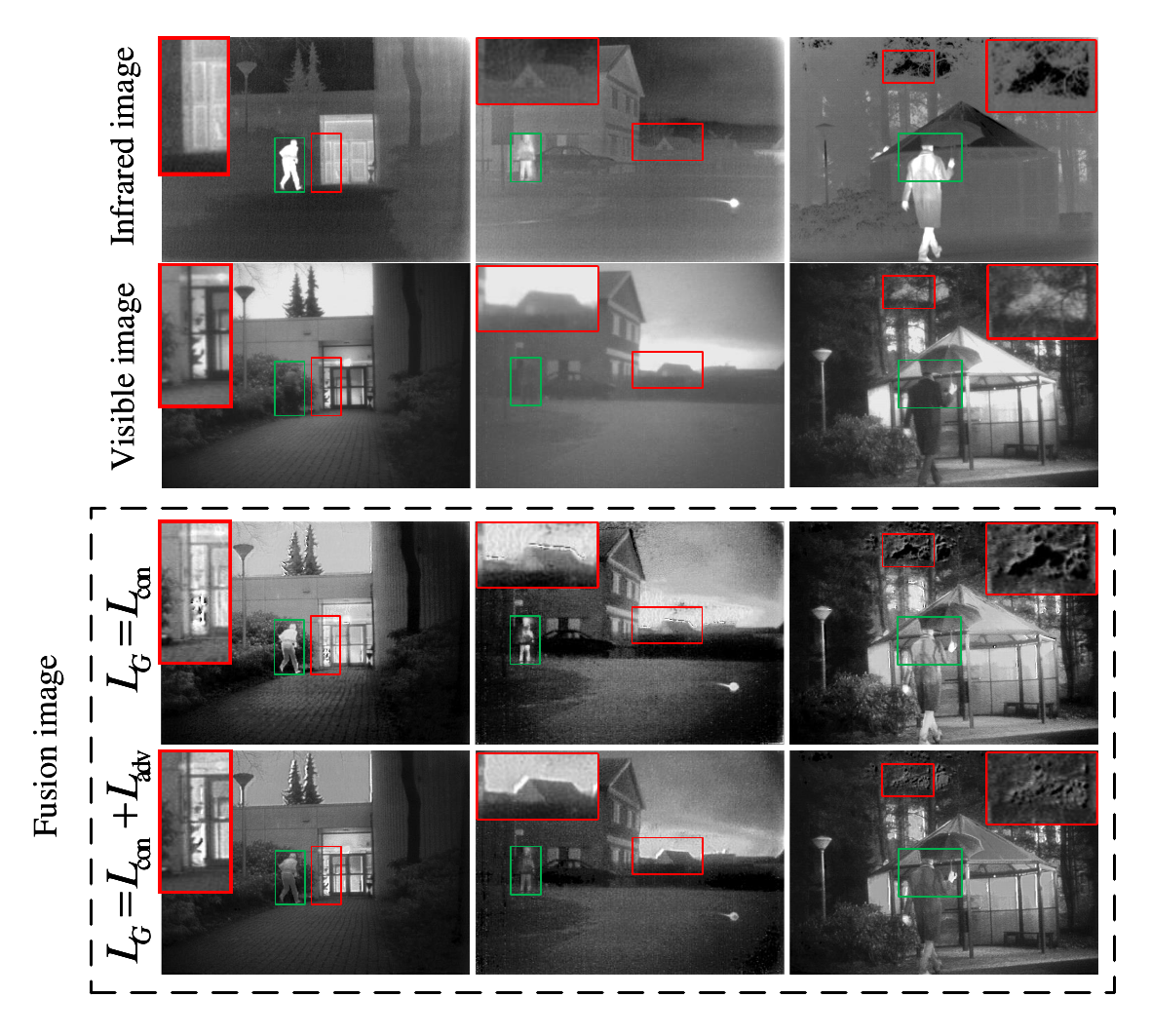}
	\caption{Fused results when the loss function of the generator $L_G$ changes. $L_{con}$ stands for the content loss and $L_{adv}$ stands for the adversarial loss.}
	\label{fig_7}
\end{figure}

The loss function of generator $L_G$ consists of two terms, \emph{i.e.}, the content loss $L_{con}$ and the adversarial loss $L_{adv}$. $L_{con}$ is designed to force the fusion image to retain the basic information from the source images. The fusion image obtained by $L_{con}$ alone are coarse, and a lot of valid information is lost. Therefore, we adopt $L_{adv}$ to encourage generator to focus on more key modality features, such as the details of visible background. 

To verify the effectiveness of $L_{con}$ and $L_{adv}$, a ablation experiment is performed. In the experiment, we remove the effect of the discriminators on the generator, and the generator is constrained by only content loss $L_{con}$. The comparative results are shown in Fig. \ref{fig_7}, where the results of the third row are determined by $L_{con}$ alone and the fourth row are determined by the combination of $L_{con}$ and $L_{adv}$. In the ablation method, $\lambda$ and $\gamma$ are set to 1 and 1, and in our method, $\lambda$, $\gamma$, $\alpha$ and $\beta$ are set to 1, 1, 10, 10 respectively. The rest of the experimental settings remain the same.

It can be seen that the results with $L_{con}$ alone have the significant target (red box) but lack enough texture details (green box), while the results with $L_{con}$ and $L_{adv}$ have more visible valid features. Obviously, content loss $L_{con}$ lacks the attention to visible structure information. Without $L_{adv}$, the visible background is susceptible to the infrared content.

\subsection{Ablation analysis of spatial domain discriminatory strategy}

\begin{figure}[H]
	\centering
	\includegraphics[scale=1]{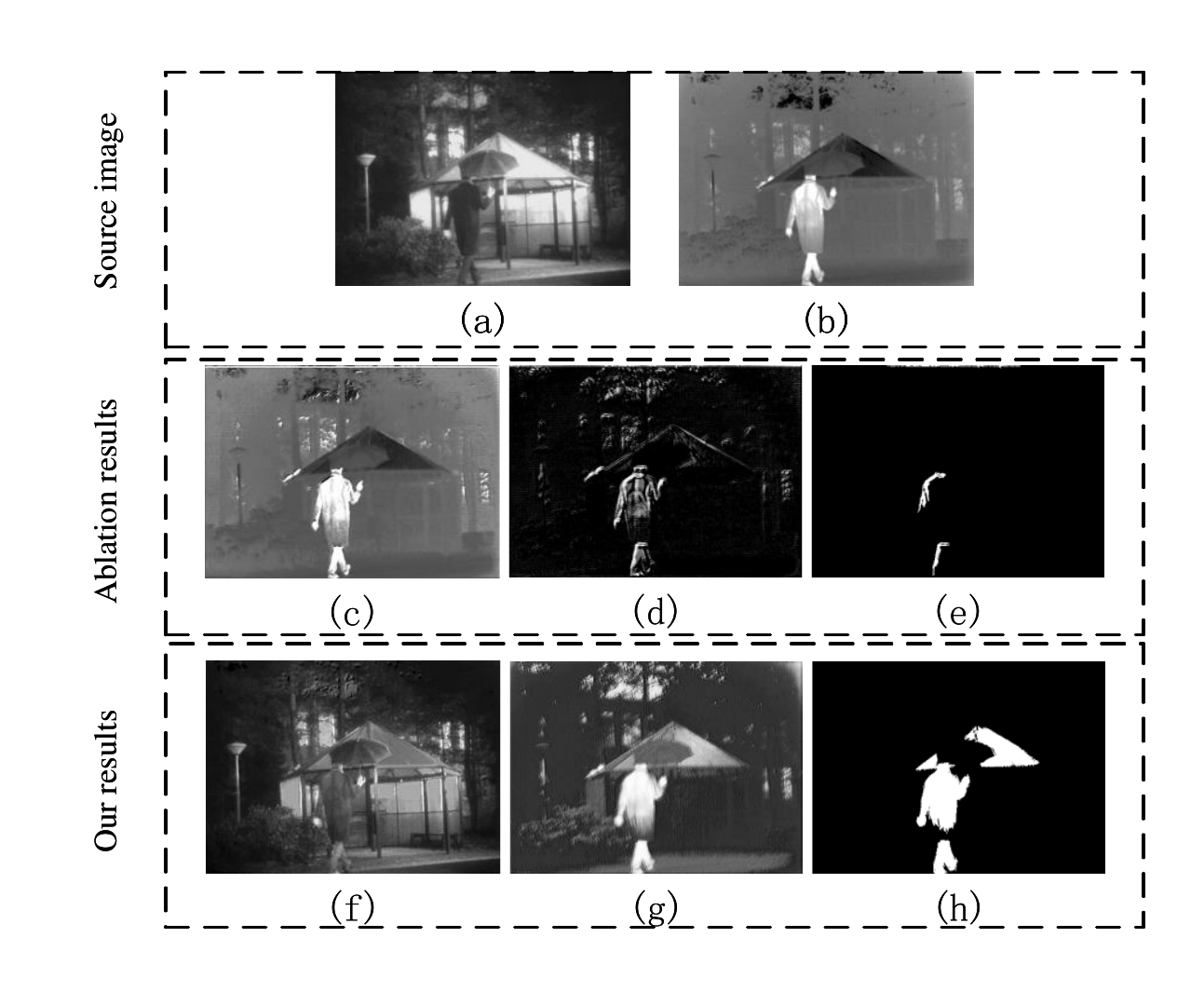}
	\caption{Results of ablation experiments based on spatial domain discriminator: Source image: (a) visible image, (b) infrared image. Ablation result: (c) fusion image, (d) infrared attention map, (e) target mask. Our result: (f) fusion image, (g) infrared attention map, (h) target mask.}
	\label{fig_8}
\end{figure}

In AWFGAN, we adopt a spatial domain discriminator $D_{spa}$ to force the generator to retain infrared information. The existing GAN-based fusion methods usually adopt the whole infrared image as the target data of the discriminator. Due to the generator prefers to focus on the smooth infrared contents as opposed to complex visible details, the discrimination based on the whole infrared image amplifies the influence of infrared information on the fusion results. To solve this problem, we design a novel spatial domain discrimination strategy with attention guidance, which can eliminate the interference of infrared background information in the discrimination process. To verify the effectiveness of the proposed discrimination strategy, we perform the ablation experiment. In the experiment, we remove the attention guidance of the spatial domain discriminator and use the original infrared image and fusion image as the input of $D_{spa}$. In the ablation method and our method, $\lambda$, $\gamma$, $\alpha$ and $\beta$ are set to 1, 1, 10, 10 respectively and the rest of the experimental settings remains the same. The results are shown in Fig. \ref{fig_8}.

As shown in Fig. \ref{fig_8}, without the guidance of attention maps, the fusion image fails to express the visible background and bias towards the infrared image. Since the discriminator targets the spatial distribution of the entire infrared image, the attention map of ablation result fails to distinguish the target region from the whole infrared scene. As a contrast, the result of our method preserves the perfect visible background and the attention map can effectively highlight the target region. It can be concluded that the spatial domain discrimination strategy with attention guidance can eliminate the interference of infrared contents on the visible background. In addition, the discriminator also acts back on the attention map, making it more accurate in extracting target regions.

\subsection{Ablation analysis of frequency domain discriminatory strategy}

\begin{figure}[H]
	\centering
	\includegraphics[scale=1]{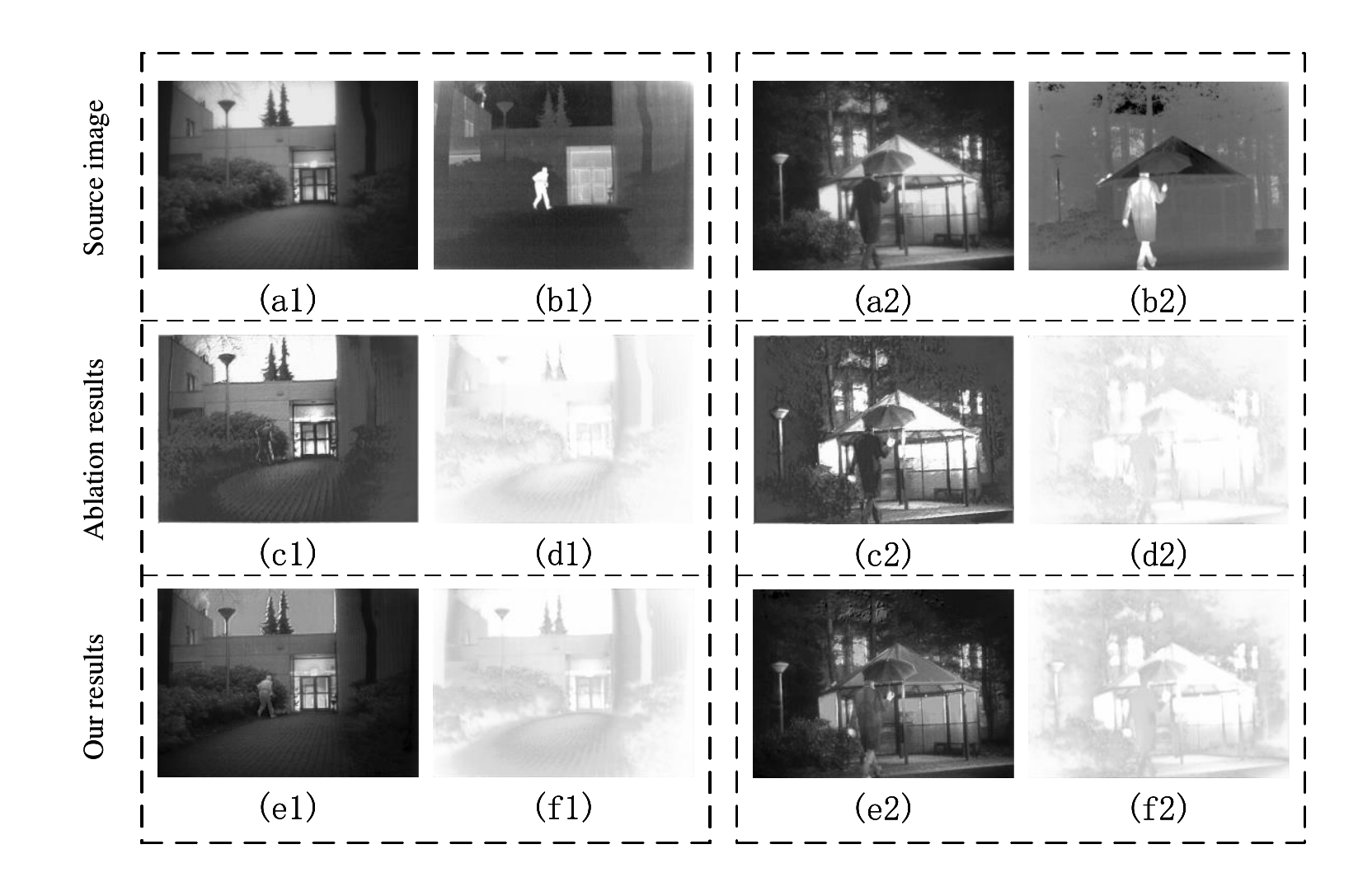}
	\caption{Results of ablation experiments based on frequency domain discriminator: Source image: (a1/a2) visible image, (b1/b2) infrared image. Ablation result: (c1/c2) fusion image, (d1/d2) visible attention map, Our result: (e1/e2) fusion image, (f1/f2) visible attention map.}
	\label{fig_9}
\end{figure}

The visible image has a large amount of structural information, which is present in the high frequency component of the visible image. In the existing GAN-based fusion method, only the pixel distribution of visible images is considered by discriminator, which tends to cause the loss of visible high frequency details. In addition, the visible discriminator has different objectives to infrared discriminator, which can enhance the confrontation between the discriminators if both discriminators use the same discriminatory strategy. To solve the above problem, we design a wavelet transform-based frequency domain discrimination strategy. We extend the range of visible information discrimination to the frequency domain based on Haar wavelets. And we design a frequency domain discriminator $D_{fre}$ which could focus on important frequency bands adaptively. With the constraint of $D_{fre}$, the generator can be forced to focus on the important visible high frequency information. 

To verify the effectiveness of the frequency domain discrimination strategy, we perform the ablation experiment. In the experiment, we replace the frequency domain discriminator with the spatial domain discriminator, and use the original visible image as the discriminatory target. In the ablation method and our method, $\lambda$, $\gamma$, $\alpha$ and $\beta$ are set to 1, 1, 10, 10 respectively and the rest of the experimental settings remains the same. The results are shown in Fig. \ref{fig_9}. 

As shown in Fig. \ref{fig_9}, the fusion images of ablation results lack the parts of visible details and significant target. And the visible attention maps lost the certain edge information. As a contrast, the result of AWFGAN retain more visible texture details. It can be seen that the frequency domain discrimination strategy weakens the confrontation between two discriminators and facilitates the retention of more infrared target contents and visible background details. Moreover, the frequency domain discriminator $D_{fre}$ motivates the SAM to focus on more edge information.

\subsection{Comparison Experiment and Analysis}

We compare AWFGAN with eight fusion methods on three public datasets. These methods include two traditional methods (ADF \cite{ADF}, FPDE \cite{FPDE}), two CNN-based methods (Densefuse \cite{li_densefuse_2019}, RFN-NEST \cite{li_rfn-nest_2021}), and four GAN-based methods (FusionGAN \cite{ma_fusiongan_2019}, GANMcC \cite{ma_ganmcc_2021}, MgAN-Fuse \cite{li_multigrained_2021} and DDcGAN \cite{ma_ddcgan_2020}). In our method, $\lambda$, $\gamma$, $\alpha$ and $\beta$ are set to 1, 1, 10, 10 respectively.

\begin{figure}[htbp]
	\centering
	\includegraphics[width=\textwidth]{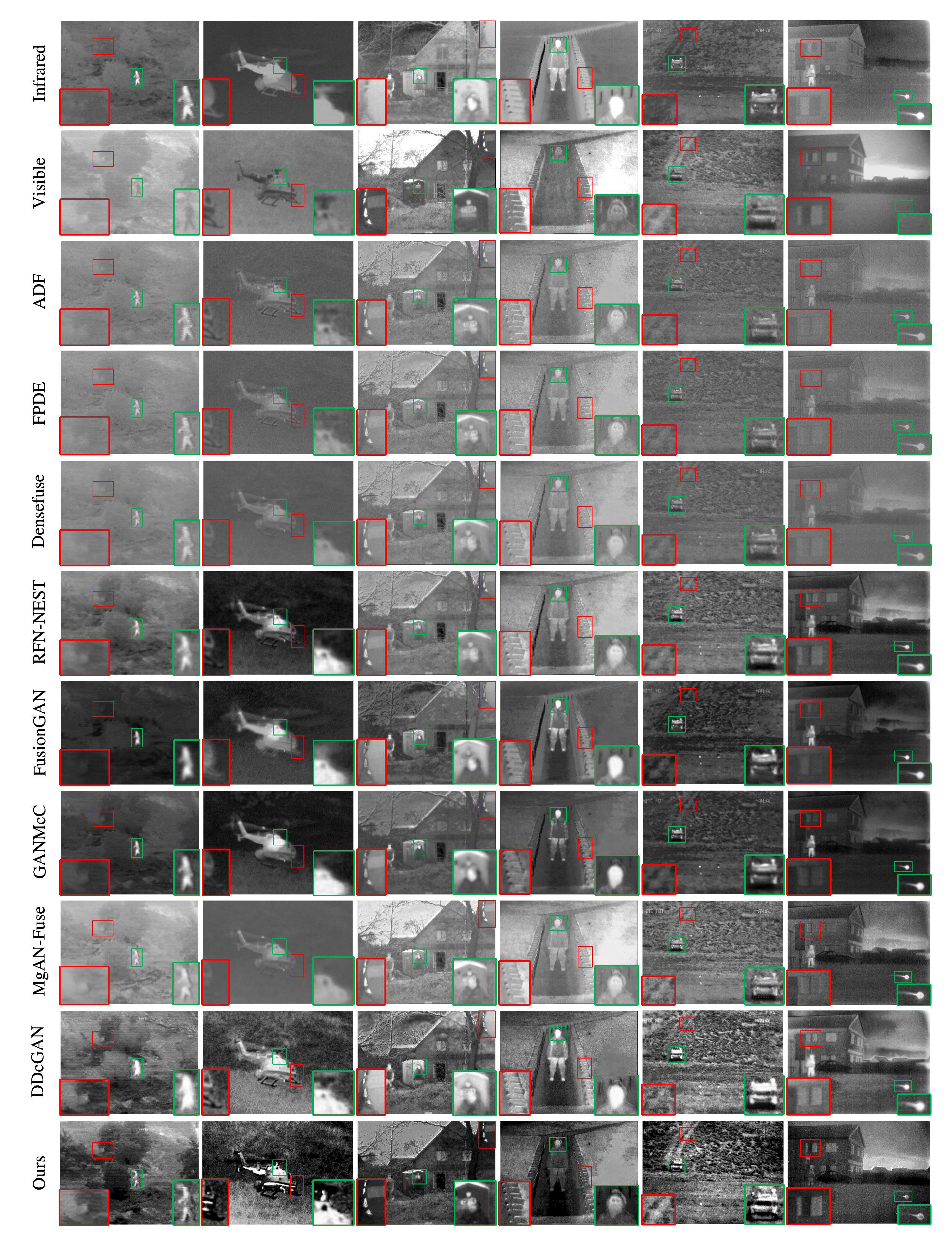}
	\caption{Qualitative comparisons of our method with eight state-of-the-art methods on TNO dataset. The region marked by green boxes represents the background detail information, and the region marked by red boxes represents the key target information.}
	\label{fig_10}
\end{figure}

\begin{figure}[!t]
	\centering
	\includegraphics[width=\textwidth]{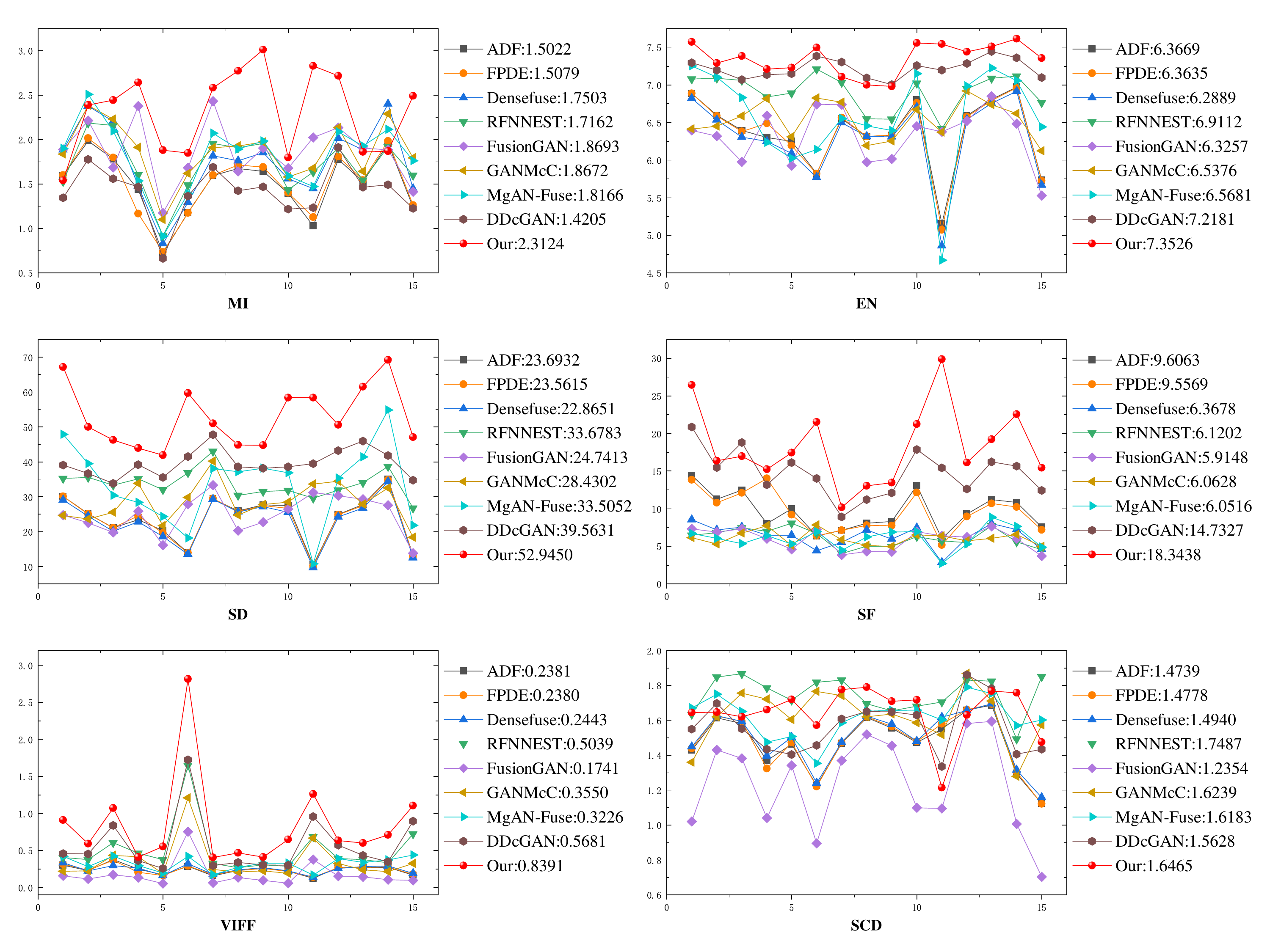}
	\caption{Quantitative comparisons of six metrics, \emph{i.e.}, MI, EN, SD, SF, VIFF and SCD, which is performed on the 15 pairs infrared and visible images of TNO dataset.}
	\label{fig_11}
\end{figure}

\begin{figure}[htbp]
	\centering
	\includegraphics[width=\textwidth]{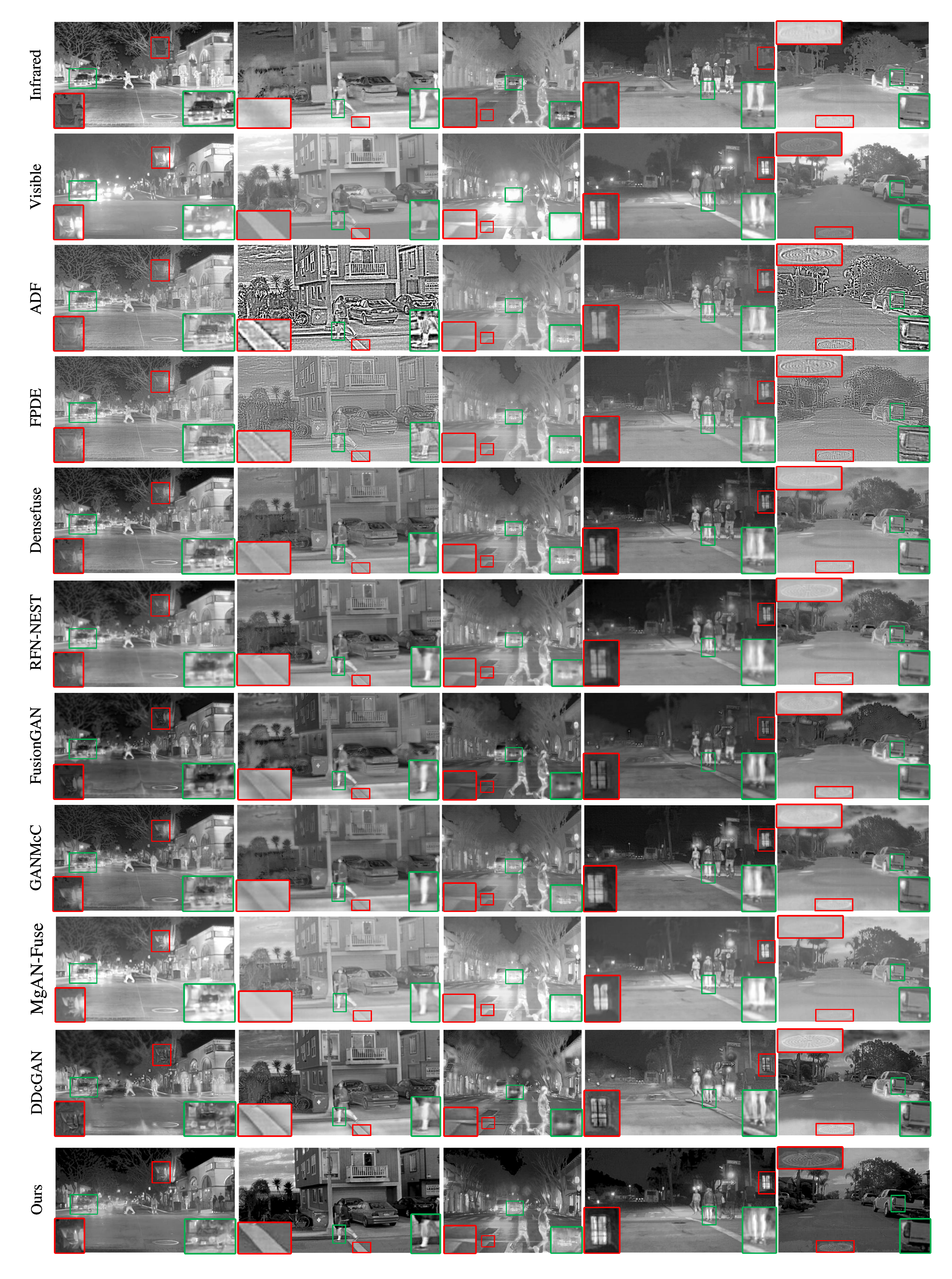}
	\caption{Qualitative comparisons of our method with eight state-of-the-art methods on RoadScene dataset. The region marked by green boxes represents the background detail information, and the region marked by red boxes represents the key target information.}
	\label{fig_12}
\end{figure}

\begin{figure}[!t]
	\centering
	\includegraphics[width=\textwidth]{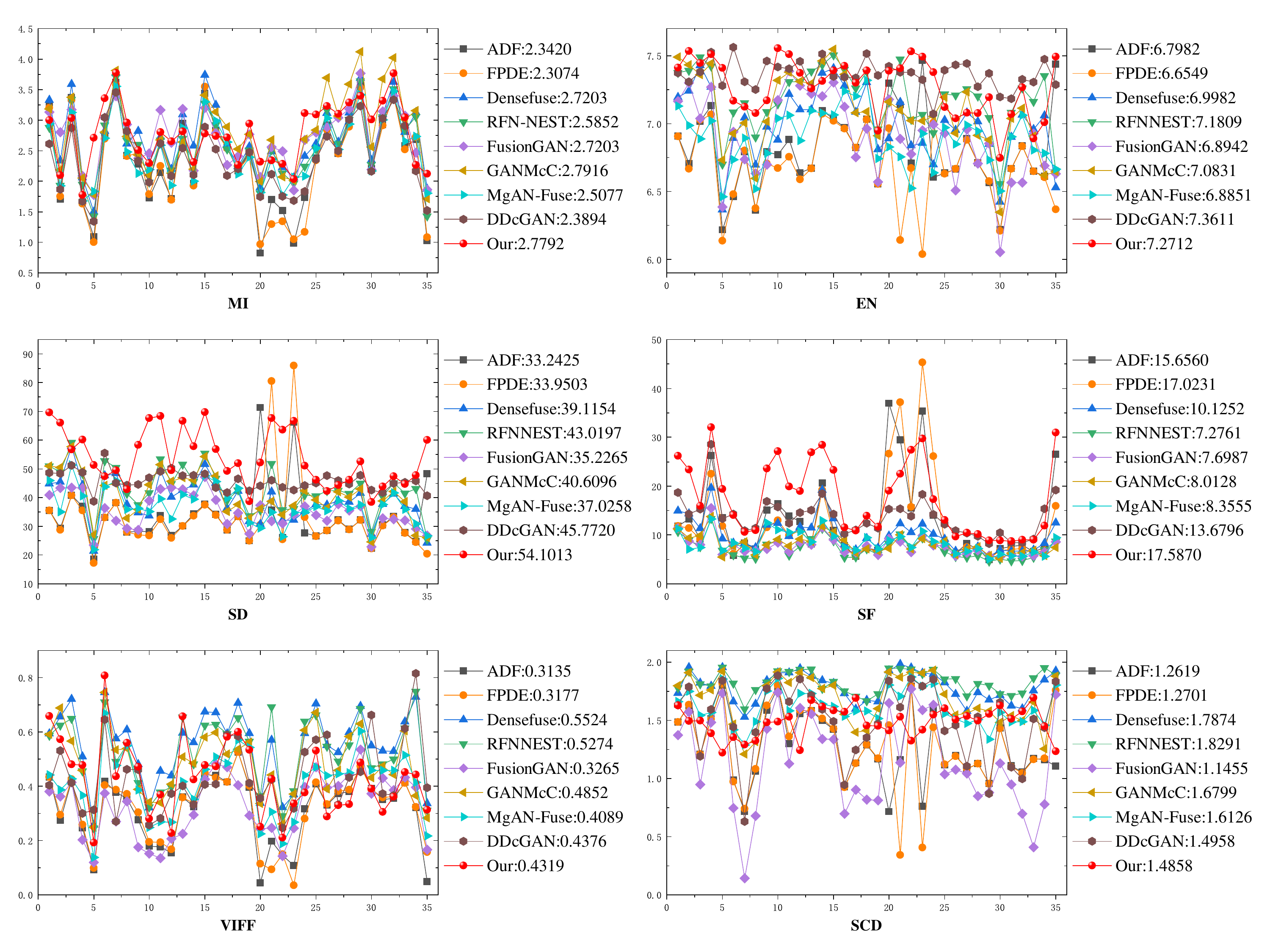}
	\caption{Quantitative comparisons of six metrics, \emph{i.e.}, MI, EN, SD, SF, VIFF and SCD, which is performed on the 35 pairs infrared and visible images of RoadScene dataset.}
	\label{fig_13}
\end{figure}

\begin{figure}[htbp]
	\centering
	\includegraphics[width=\textwidth]{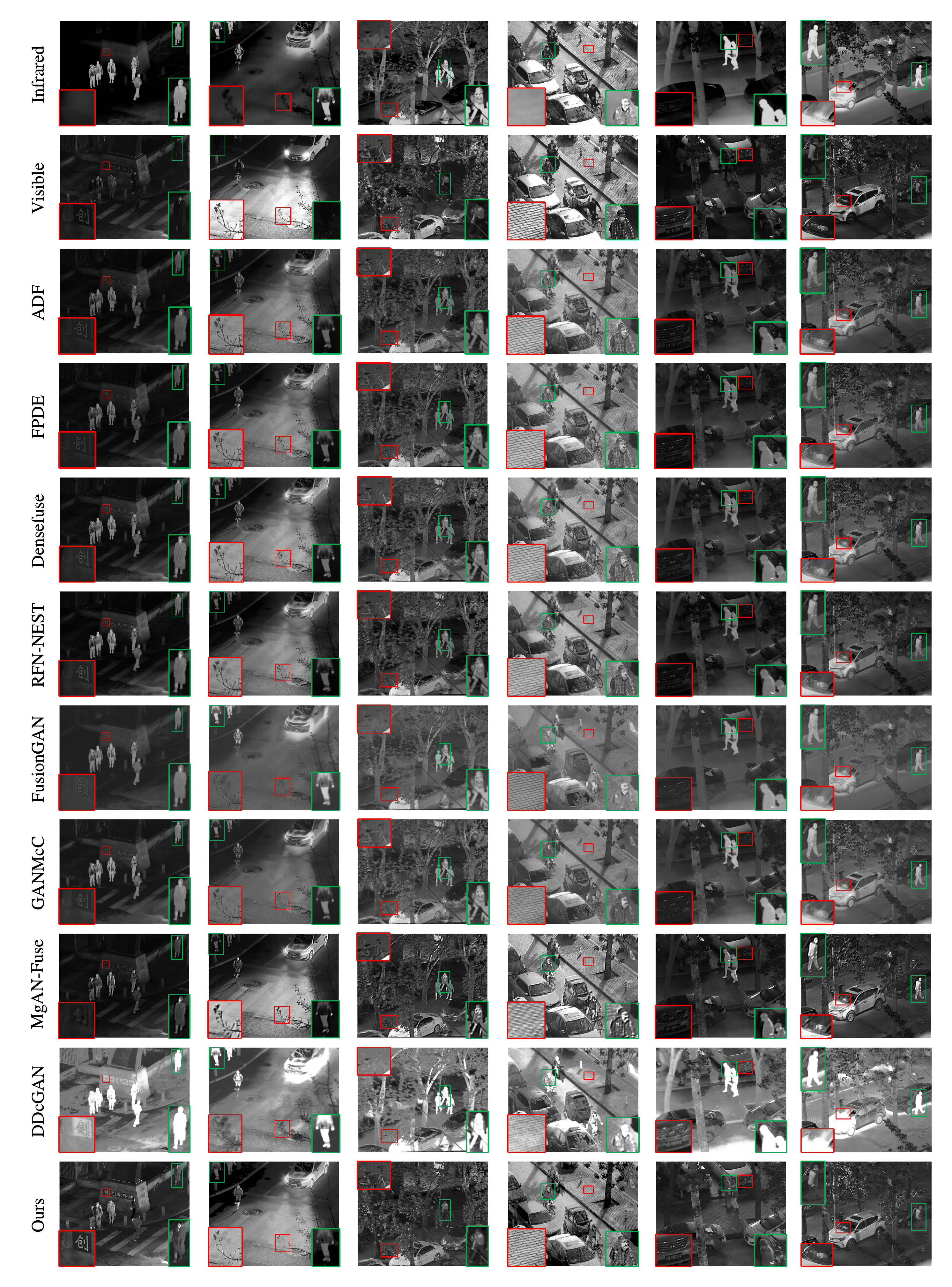}
	\caption{Qualitative comparisons of our method with eight state-of-the-art methods on LLVIP dataset. The region marked by green boxes represents the background detail information, and the region marked by red boxes represents the key target information.}
	\label{fig_14}
\end{figure}

\begin{figure}[!t]
	\centering
	\includegraphics[width=\textwidth]{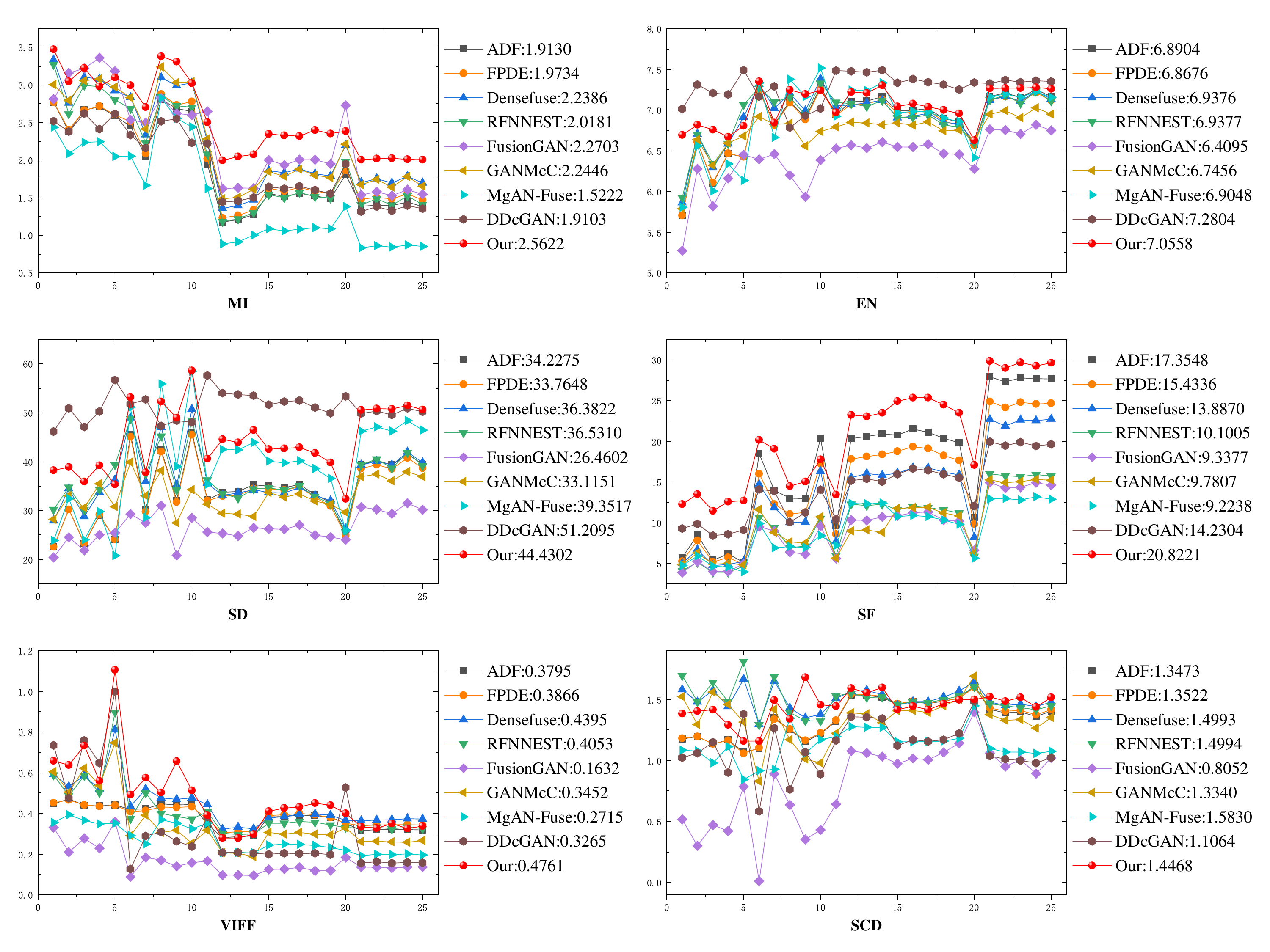}
	\caption{Quantitative comparisons of six metrics, \emph{i.e.}, MI, EN, SD, SF, VIFF and SCD, which is performed on the 25 pairs infrared and visible images of LLVIP dataset.}
	\label{fig_15}
\end{figure}

(1) Comparison on TNO dataset: we test nine methods on the TNO dataset, and some results are shown in Fig. \ref{fig_10}. The first and second rows are source images, and the third to eleventh rows correspond to the results of ADF, FPDE, Densefuse, RFN-NEST, FusionGAN, GANMcC, MgAN-Fuse, DDcGAN and AWFGAN in turn.

In order to better describe the fusion performance, we select some representative regions which contain rich backgrounds or key targets. The regions marked by green boxes represent the background detail information, and the regions marked by red boxes represent the key target information. By comparison, it can be seen that the results of traditional and CNN-based fusion methods are relatively smooth. For instance, in the fusion image of ADF, FPDE, Densefuse and RFN-NEST, the sky is inclined to gray, close to the brightness of the ground. In addition, part of visible background is broken by the infrared contents in these fusion results, for example, in the first and fifth columns in the second column, the ground textures of ADF and FPDE are lost; the helicopter of Densefuse retain mainly the infrared features. In the GAN-based results, the FusionGAN and GANMcC fail to retain perfect visible background. The DDcGAN and MgAN-Fuse have the superior fusion performance, but the visible details of the infrared target is not preserved enough. Our results have the balanced fusion performance, which are close to visible images in the background regions, and close to infrared image in the target regions. The visible details are sufficiently preserved in our results, such as the edges of the road and the street sign. Moreover, our results add the visible details to the infrared targets while expressing them significantly. For example, in the fourth column, our results preserve the visible features of the human face.

Quantitative analysis is performed on 15 pairs of images of the TNO dataset, and the results are shown in Fig. \ref{fig_11}. The comparison shows that our results have significant advantage on MI, EN, SD, SF and VIFF, and are second only to RFN-NEST on SCD. From the results, we can conclude the following points. Firstly, the superior performances on MI and EN demonstrate that AWFGAN preserves a large amount of valid information, and it achieves a better balance between two types of modality information. By comparing the values of SD, it can be concluded that AWFGAN can generate the fusion results with the high contrast, which means that our results have a good visual effect. In the meantime, the largest SF demonstrate that our method preserves sufficient texture details and achieves better image quality. Last but not least, the advantage of VIFF and SCD illustrat the ability of our method to completely retain the visual and valid information of the source image.

(2) Comparison on RoadScene dataset: we test nine methods on the RoadScene dataset, the partial results are shown in Fig. \ref{fig_12}. The first and second rows are source images, and the third to eleventh rows correspond to the results of ADF, FPDE, Densefuse, RFN-NEST, FusionGAN, GANMcC, MgAN-Fuse, DDcGAN and AWFGAN in turn.

Fig. \ref{fig_12} shows parts of typical results of RoadScene dataset. By comparison, it can be seen that AWFGAN preserves more visible background details, for instance, the advertisements of first column, the crosswalks of second and third columns, windows of forth column and manhole cover of fifth column. In addition, more target details are retained in our results, such as the details of the car hidden by the lights of first and third columns, the clothes of the pedestrians of second and forth columns and the logo of the car of fifth column.

Quantitative analysis is performed on 35 pairs of images of the RoadScene dataset, which is shown in Fig. \ref{fig_13}. Similar to the performance on the TNO dataset, our results have superior performance on the values of SD and SF. Moreover, our results are second only to GANMcC on MI and second only to DDcGAN on EN. The above demonstrates the superiority of our method in terms of image quality and effective information retention. The VIFF and SCD of our results are not optimal and rank in the middle, indicating the poor generalization of our method on the RoadScene dataset, but the visual effect in the subjective comparison shows that our results are acceptable.

(3) Comparison on LLVIP dataset: we also test nine methods on the LLVIP dataset, the partial results are shown in Fig. \ref{fig_14}. The first and second rows are source images, and the third to eleventh rows correspond to the results of ADF, FPDE, Densefuse, RFN-NEST, FusionGAN, GANMcC, MgAN-Fuse, DDcGAN and AWFGAN in turn.

Compared to the results on the TNO and RoadScene datasets, our fusion images on LLVIP are able to more fully retain visible detail information, such as the text in the first and fifth columns, tree branch textures in the second and third columns, and ground features in the fourth column and car details in the sixth column. In addition, parts of the contents of infrared target are also supplemented by visible information, for example, the pedestrians in all images retain more complete visible details.

Quantitative analysis is performed on 25 pairs of images of the LLVIP dataset, which is shown in Fig. \ref{fig_15}. On MI, our method can achieve the best quantization result, which illustrates the superiority of AWFGAN on information retention. On EN and SD, our method is second only to DDcGAN. However, the visual effect of DDcGAN is not enough, the reason is that DDcGAN over-expresses infrared information, which causes a strong contrast between the background and the target, but this also introduces a certain amount of noise, which causes damage to the original information from the source images. Moreover, we have the best SF and VIFF, but the SCD is not as good as Densefuse, RFN-NEST and MgAN-Fuse, indicating that our method can greatly preserve the visual information of the original image.

\subsection{Efficiency Comparison}

To verify the fusion efficiency of different methods, we counted the average running time of nine methods on three datasets, and the comparison results are shown in Table 1. It is observed that the traditional methods have a clear advantage on TNO and RoadScene, while the deep learning methods are more efficient on LLVIP with larger pixels. Our method has the highest efficiency on the three datasets due to the simple generator structure.

\begin{table}[H]
	\caption{Mean values of the running times of all methods on the TNO, RoadScene and LLVIP datasets (unit: second).}
	\centering
	\setlength{\tabcolsep}{9mm}{
	\begin{tabular}{lccc}
		\toprule
		Methods&TNO&RoadScene&LLVIP \\
		\midrule
		ADF&0.1019&0.9341&6.4939   \\
		FPDE&0.2666&1.8034&19.9443 \\
		Densefuse&0.1941&4.0016&2.3342   \\
		RFNNEST&0.2472&1.8962&17.3514  \\
		FusionGAN&0.3152&6.4940&7.2457  \\
		GANMcC&0.6027&12.2204&14.1148  \\
		MgAN-Fuse&0.1677&1.2792&12.1877  \\
		DDcGAN&0.7759&15.6278&14.3230  \\
		Ours&0.0587&0.5002&4.2208  \\
		\bottomrule
	\end{tabular}}
\end{table}

\section{Conclution}
In this article, we propose a novel GAN-based image fusion method (AWFGAN), which adopt two unique discrimination strategies to improve the fusion performance. We first introduce the spatial attention modules (SAM) into the generator to obtain the spatial attention maps. Subsequently the attention maps are utilized to limit the discrimination of infrared images to the target regions. This strategy effectively reduces the impact of infrared contents on the fused background. In addition, we extend the discrimination range of visible information to the wavelet subspace, which can force the generator to restore the high-frequency details of visible images. We validate the effectiveness and superior of AWFGAN on three public datasets (TNO, RoadScene and LLVIP). As a result, AWFGAN can preserve sufficient texture details in the background regions and meets the visual requirements in the target regions. Our method also adds a certain visible information to the target regions, which helps in the processing of subsequent advanced vision tasks. The extensive ablation and comparison experiments also demonstrate that our method have superior fusion performance.

\section*{Acknowledgments}
This work was supported by the Program of High Resolution Earth Observation under Grant 01−Y30F05−9001−20/22 and GFZX0404130307, the Key Research and Transformation project of Qinghai Province 2022-SF-150, the National Natural Science Foundation of China under Grant 62206321.


\begin{thebibliography}{10}
	\expandafter\ifx\csname url\endcsname\relax
	\def\url#1{\texttt{#1}}\fi
	\expandafter\ifx\csname urlprefix\endcsname\relax\def\urlprefix{URL }\fi
	\expandafter\ifx\csname href\endcsname\relax
	\def\href#1#2{#2} \def\path#1{#1}\fi
	
	\bibitem{li_pixel-level_2017}
	S.~Li, X.~Kang, L.~Fang, J.~Hu, H.~Yin, Pixel-level image fusion: {A} survey of
	the state of the art, Information Fusion 33 (2017) 100--112.
	
	\bibitem{ma_infrared_2019}
	J.~Ma, Y.~Ma, C.~Li, Infrared and visible image fusion methods and
	applications: {A} survey, Information Fusion 45 (2019) 153--178.
	
	\bibitem{rajah_feature_2018}
	P.~Rajah, J.~Odindi, O.~Mutanga, Feature level image fusion of optical imagery
	and {Synthetic} {Aperture} {Radar} ({SAR}) for invasive alien plant species
	detection and mapping, Remote Sensing Applications: Society and Environment
	10 (2018) 198--208.
	
	\bibitem{chan_fusing_2013}
	A.~L. Chan, S.~R. Schnelle, Fusing concurrent visible and infrared videos for
	improved tracking performance, Optical Engineering 52~(1) (2013) 017004.
	
	\bibitem{han_fusion_2007}
	J.~Han, B.~Bhanu, Fusion of color and infrared video for moving human
	detection, Pattern Recognition 40~(6) (2007) 1771--1784.
	
	\bibitem{singh_integrated_2008}
	R.~Singh, M.~Vatsa, A.~Noore, Integrated multilevel image fusion and match
	score fusion of visible and infrared face images for robust face recognition,
	Pattern Recognition 41~(3) (2008) 880--893.
	
	\bibitem{dogra_multi-scale_2017}
	A.~Dogra, B.~Goyal, S.~Agrawal, From {Multi}-{Scale} {Decomposition} to
	{Non}-{Multi}-{Scale} {Decomposition} {Methods}: {A} {Comprehensive} {Survey}
	of {Image} {Fusion} {Techniques} and {Its} {Applications}, IEEE Access 5
	(2017) 16040--16067.
	
	\bibitem{ADF}
	D.~P. Bavirisetti, R.~Dhuli, Fusion of {Infrared} and {Visible} {Sensor}
	{Images} {Based} on {Anisotropic} {Diffusion} and {Karhunen}-{Loeve}
	{Transform}, IEEE Sensors Journal 16~(1) (2016) 203--209.
	
	\bibitem{FPDE}
	D.~P. Bavirisetti, G.~Xiao, G.~Liu, Multi-sensor image fusion based on fourth
	order partial differential equations, 2017 20th International Conference on
	Information Fusion (Fusion) (2017) 1--9.
	
	\bibitem{liu_fusion_2017}
	Z.~Liu, Y.~Feng, H.~Chen, L.~Jiao, A fusion algorithm for infrared and visible
	based on guided filtering and phase congruency in {NSST} domain, Optics and
	Lasers in Engineering 97 (2017) 71--77.
	
	\bibitem{olshausen_emergence_1996}
	B.~A. Olshausen, D.~J. Field, Emergence of simple-cell receptive field
	properties by learning a sparse code for natural images, Nature 381~(6583)
	(1996) 607--609.
	
	\bibitem{zhou_principal_2011}
	Y.~Zhou, A.~Mayyas, M.~Omar, Principal {Component} {Analysis}-{Based} {Image}
	{Fusion} {Routine} with {Application} to {Automotive} {Stamping} {Split}
	{Detection}, Research in Nondestructive Evaluation 22~(2) (2011) 76--91.
	
	\bibitem{kong_technique_2010}
	W.~Kong, Y.~Lei, Y.~Lei, J.~Zhang, Technique for image fusion based on
	non-subsampled contourlet transform domain improved {NMF}, Science China
	Information Sciences 53~(12) (2010) 2429--2440.
	
	\bibitem{cvejic_region-based_2007}
	N.~Cvejic, D.~Bull, N.~Canagarajah, Region-{Based} {Multimodal} {Image}
	{Fusion} {Using} {ICA} {Bases}, IEEE Sensors Journal 7~(5) (2007) 743--751.
	
	\bibitem{sampat_visible_2015}
	T.~Shibata, M.~Tanaka, M.~Okutomi, Visible and near-infrared image fusion based
	on visually salient area selection, San Francisco, California, United States,
	2015, p. 94040G.
	
	\bibitem{li_poisson_2019}
	J.~Li, H.~Huo, C.~Sui, C.~Jiang, C.~Li, Poisson {Reconstruction}-{Based}
	{Fusion} of {Infrared} and {Visible} {Images} via {Saliency} {Detection},
	IEEE Access 7 (2019) 20676--20688.
	
	\bibitem{bavirisetti_two-scale_2016}
	D.~P. Bavirisetti, R.~Dhuli, Two-scale image fusion of visible and infrared
	images using saliency detection, Infrared Physics \& Technology 76 (2016)
	52--64.
	
	\bibitem{liu_general_2015}
	Y.~Liu, S.~Liu, Z.~Wang, A general framework for image fusion based on
	multi-scale transform and sparse representation, Information Fusion 24 (2015)
	147--164.
	
	\bibitem{chen_infrared_2022}
	Y.~Chen, L.~Cheng, H.~Wu, F.~Mo, Z.~Chen, Infrared and visible image fusion
	based on iterative differential thermal information filter, Optics and Lasers
	in Engineering 148 (2022) 106776.
	
	\bibitem{liu_deep_2018}
	Y.~Liu, X.~Chen, Z.~Wang, Z.~J. Wang, R.~K. Ward, X.~Wang, Deep learning for
	pixel-level image fusion: {Recent} advances and future prospects, Information
	Fusion 42 (2018) 158--173.
	
	\bibitem{liu_multi-focus_2017}
	Y.~Liu, X.~Chen, H.~Peng, Z.~Wang, Multi-focus image fusion with a deep
	convolutional neural network, Information Fusion 36 (2017) 191--207.
	
	\bibitem{ma_fusiongan_2019}
	J.~Ma, W.~Yu, P.~Liang, C.~Li, J.~Jiang, {FusionGAN}: {A} generative
	adversarial network for infrared and visible image fusion, Information Fusion
	48 (2019) 11--26.
	
	\bibitem{ma_ddcgan_2020}
	J.~Ma, H.~Xu, J.~Jiang, X.~Mei, X.-P. Zhang, {DDcGAN}: {A}
	{Dual}-{Discriminator} {Conditional} {Generative} {Adversarial} {Network} for
	{Multi}-{Resolution} {Image} {Fusion}, IEEE Transactions on Image Processing
	29 (2020) 4980--4995.
	
	\bibitem{li_attentionfgan_2021}
	J.~Li, H.~Huo, C.~Li, R.~Wang, Q.~Feng, {AttentionFGAN}: {Infrared} and
	{Visible} {Image} {Fusion} {Using} {Attention}-{Based} {Generative}
	{Adversarial} {Networks}, IEEE Transactions on Multimedia 23 (2021)
	1383--1396.
	
	\bibitem{heijmans_nonlinear_2000}
	H.~Heijmans, J.~Goutsias, Nonlinear multiresolution signal decomposition
	schemes. {II}. {Morphological} wavelets, IEEE Transactions on Image
	Processing 9~(11) (2000) 1897--1913.
	
	\bibitem{li_infrared_2019}
	H.~Li, X.-j. Wu, T.~S. Durrani, Infrared and visible image fusion with {ResNet}
	and zero-phase component analysis, Infrared Physics \& Technology 102 (2019)
	103039.
	
	\bibitem{li_densefuse_2019}
	H.~Li, X.-J. Wu, {DenseFuse}: {A} {Fusion} {Approach} to {Infrared} and
	{Visible} {Images}, IEEE Transactions on Image Processing 28~(5) (2019)
	2614--2623.
	
	\bibitem{zhang_ifcnn_2020}
	Y.~Zhang, Y.~Liu, P.~Sun, H.~Yan, X.~Zhao, L.~Zhang, {IFCNN}: {A} general image
	fusion framework based on convolutional neural network, Information Fusion 54
	(2020) 99--118.
	
	\bibitem{ma_stdfusionnet_2021}
	J.~Ma, L.~Tang, M.~Xu, H.~Zhang, G.~Xiao, {STDFusionNet}: {An} {Infrared} and
	{Visible} {Image} {Fusion} {Network} {Based} on {Salient} {Target}
	{Detection}, IEEE Transactions on Instrumentation and Measurement 70 (2021)
	1--13.
	
	\bibitem{jian_sedrfuse_2021}
	L.~Jian, X.~Yang, Z.~Liu, G.~Jeon, M.~Gao, D.~Chisholm, {SEDRFuse}: {A}
	{Symmetric} {Encoder}–{Decoder} {With} {Residual} {Block} {Network} for
	{Infrared} and {Visible} {Image} {Fusion}, IEEE Transactions on
	Instrumentation and Measurement 70 (2021) 1--15.
	
	\bibitem{li_rfn-nest_2021}
	H.~Li, X.-J. Wu, J.~Kittler, {RFN}-{Nest}: {An} end-to-end residual fusion
	network for infrared and visible images, Information Fusion 73 (2021) 72--86.
	
	\bibitem{li_cgtf_2022}
	J.~Li, J.~Zhu, C.~Li, X.~Chen, B.~Yang, {CGTF}: {Convolution}-{Guided}
	{Transformer} for {Infrared} and {Visible} {Image} {Fusion}, IEEE
	Transactions on Instrumentation and Measurement 71 (2022) 1--14.
	
	\bibitem{li_infrared_2020}
	J.~Li, H.~Huo, K.~Liu, C.~Li, Infrared and visible image fusion using dual
	discriminators generative adversarial networks with {Wasserstein} distance,
	Information Sciences 529 (2020) 28--41.
	
	\bibitem{ma_ganmcc_2021}
	J.~Ma, H.~Zhang, Z.~Shao, P.~Liang, H.~Xu, {GANMcC}: {A} {Generative}
	{Adversarial} {Network} {With} {Multiclassification} {Constraints} for
	{Infrared} and {Visible} {Image} {Fusion}, IEEE Transactions on
	Instrumentation and Measurement 70 (2021) 1--14.
	
	\bibitem{li_multigrained_2021}
	J.~Li, H.~Huo, C.~Li, R.~Wang, C.~Sui, Z.~Liu, Multigrained {Attention}
	{Network} for {Infrared} and {Visible} {Image} {Fusion}, IEEE Transactions on
	Instrumentation and Measurement 70 (2021) 1--12.
	
	\bibitem{YANG2022116905}
	X.~Yang, H.~Huo, J.~Li, C.~Li, Z.~Liu, X.~Chen, Dsg-fusion: Infrared and
	visible image fusion via generative adversarial networks and guided filter,
	Expert Systems with Applications 200 (2022) 116905.
	
	\bibitem{Gulrajani_improved_2017}
	I.~Gulrajani, F.~Ahmed, M.~Arjovsky, V.~Dumoulin, A.~Courville, Improved
	training of wasserstein gans, Advances in Neural Information Processing
	Systems 10 (2017) 5767--5777.
	
	\bibitem{qu_information_2002}
	G.~Qu, D.~Zhang, P.~Yan, Information measure for performance of image fusion,
	Electronics Letters 38~(7) (2002) 313.
	
	\bibitem{van_aardt_assessment_2008}
	J.~Van~Aardt, Assessment of image fusion procedures using entropy, image
	quality, and multispectral classification, Journal of Applied Remote Sensing
	2~(1) (2008) 023522.
	
	\bibitem{rao_-fibre_1997}
	Y.-J. Rao, In-fibre {Bragg} grating sensors, Measurement Science and Technology
	8~(4) (1997) 355--375.
	
	\bibitem{eskicioglu_image_1995}
	A.~Eskicioglu, P.~Fisher, Image quality measures and their performance, IEEE
	Transactions on Communications 43~(12) (1995) 2959--2965.
	
	\bibitem{VIFF}
	Y.~Han, Y.~Cai, Y.~Cao, X.~Xu, A new image fusion performance metric based on
	visual information fidelity, Inf. Fusion 14 (2013) 127--135.
	
	\bibitem{SCD}
	V.~Aslantas, E.~Bendes, A new image quality metric for image fusion: The sum of
	the correlations of differences, AEU - International Journal of Electronics
	and Communications 69~(12) (2015) 1890--1896.
	
\end{thebibliography}
\end{document}